\documentclass{article}
\usepackage[final, nonatbib]{neurips_2024}
\usepackage{graphicx} 
\usepackage{float}
\usepackage{tabularx}
\usepackage{adjustbox}
\usepackage{appendix}
\usepackage{booktabs}
\usepackage{enumitem}
\usepackage{subcaption}
\usepackage{verbatim}
\usepackage{colortbl}
\usepackage{multirow}
\usepackage{dsfont}
\usepackage[style=apa]{biblatex}
\usepackage{hyperref}
\usepackage{amsmath}
\usepackage{amssymb}
\usepackage{stmaryrd}
\hypersetup{
    colorlinks=true, 
    linkcolor=blue,  
    citecolor=blue,  
    urlcolor=blue    
}

\DeclareCiteCommand{\cite}
  {\usebibmacro{prenote}}
  {\usebibmacro{citeindex}%
   \printtext[bibhyperref]{(\usebibmacro{cite})}}
  {\multicitedelim}
  {\usebibmacro{postnote}}

\title{Foundation models for time series forecasting: Application in conformal prediction}
\author{%
  Sami Achour\textsuperscript{1},
  Yassine Bouher\textsuperscript{2}\thanks{Work done during his internship at Ekimetrics.} ,
  Duong Nguyen\textsuperscript{1},
  and Nicolas Chesneau\textsuperscript{1}\\
  \textsuperscript{1}Ekimetrics,
  \textsuperscript{2}École Polytechnique\\
  \texttt{firstname.lastname@ekimetrics.com, yassine.bouher@polytechnique.edu}
}
\date{Date}
\begin{document}

\maketitle

\begin{abstract}


The zero-shot capabilities of foundation models (FMs) for time series forecasting offer promising potentials in conformal prediction, as most of the available data can be allocated to calibration. This study compares the performance of Time Series Foundation Models (TSFMs) with traditional methods, including statistical models and gradient boosting, within a conformal prediction setting. Our findings highlight two key advantages of TSFMs. First, when the volume of data is limited, TSFMs provide more reliable conformalized prediction intervals than classic models, thanks to their superior predictive accuracy. Second, the calibration process is more stable because more data are used for calibration. Morever, the fewer data available, the more pronounced these benefits become, as classic models require a substantial amount of data for effective training. These results underscore the potential of foundation models in improving conformal prediction reliability in time series applications, particularly in data-constrained cases. All the code to reproduce the experiments is available on \href{https://github.com/ekimetrics/ts-genai-conformal}{GitHub}.
    
\end{abstract}

\section{Introduction}
\label{sec:introduction}

Accurate forecasting of time series is crucial for informed decision-making and strategic planning in various domains, such as finance \cite{lee2021learning}, healthcare \cite{parker2024interactive}, climate science \cite{article}, and retails \cite{carbonneau2008application}. Machine learning approaches, ranging from classical statistical methods (e.g., ARIMA \cite{box1970time}, exponential smoothing \cite{hyndman2021forecasting}) to tree-based models (e.g., XGBoost \cite{chen2016xgboost}, LightGBM \cite{ke2017lightgbm}, CatBoost \cite{dorogush2018catboost}) and modern deep learning architectures (e.g.,  DeepAR \cite{salinas2019deeparprobabilisticforecastingautoregressive}, Autoformer \cite{wu2022autoformerdecompositiontransformersautocorrelation}, FEDformer \cite{zhou2022fedformerfrequencyenhanceddecomposed}, TFT \cite{lim2020temporalfusiontransformersinterpretable}, and PatchTST \cite{nie2023timeseriesworth64}) have been widely used to tackle this challenge \cite{kolassa2023demand}.

Recently, the emergence of foundation models (FMs) for time series presents new opportunities for enhancing forecast accuracy and reliability \cite[]{aksu2024gift, liang2024foundation}.
Foundation models, characterized by the pre-training of large-scale models on extensive and diverse datasets, are designed to learn a wide range of patterns and representations, which can then be applied, fine-tuned or adapted for specific downstream tasks across various domains with zero or minimal additional training. The foundation models approach utilizes the broad knowledge gained during pre-training, allowing models to generalize well and perform tasks in a zero-shot \cite{xian2018zero} or few-shot \cite{song2022comprehensivesurveyfewshotlearning} mode, which reduces the need for large volumes of task-specific data. Foundation models have achieved remarkable success in Natural Language Processing (NLP) \cite{Khurana_2022}, Computer Vision (CV) \cite{CHAI2021100134}, and are increasingly being explored for its potential in other areas, including time series modeling \cite[]{liang2024foundation}.
Several recent efforts have yielded promising results. Notable examples include Nixtla's TimeGPT \cite{garza2024timegpt1}, Amazon's Chronos \cite{ansari2024chronoslearninglanguagetime}, or Google's TimeFM \cite{das2024decoderonlyfoundationmodeltimeseries} have achieved forecasting accuracy comparable to, or in some cases exceeding, state-of-the-art statistical and machine learning methods \cite[]{aksu2024gift}.
The potential of Time Series Foundation Models (TSFMs) is considerable, offering opportunities to reduce predictive errors, enhance model generalisation across diverse time series datasets, and facilitate the integration of multi-modal data to address complex predictive and analytical tasks.

In this article, we explore a different aspect of their potential: FMs for time series forecasting in the context of Conformal Prediction (CP).
Conformal prediction \cite{shafer2007tutorialconformalprediction} is a robust statistical method for uncertainty quantification that provides predictive intervals with a specified confidence level. CP constructs intervals that are expected to contain the true value in a proportion of cases equal to the chosen confidence level, e.g. 90\% of the time. To achieve this, CP requires a calibration set—a separate portion of data distinct from the training set. The method first trains a model on the training data, then applies it to the calibration set to assess how well the model’s predictions align with actual outcomes. This alignment enables CP to adjust the interval widths accordingly. For a comprehensive introduction, readers are referred to \cite{gentleintroductiontocp}. While CP is effective for providing reliable uncertainty estimates, the need for a calibration set can be a limitation when data is scarce, as it requires partitioning an already small dataset into training and calibration subsets, which may impact model performance and calibration accuracy.
The zero shot prediction capabilities of FMs make them particularly appealing for CP in time series forecasting, as training is no longer a constraint, enabling most available data to be allocated for calibration. In this article, we explore this approach by applying CP to open-source TSFMs on public datasets. Specifically, we tested three TSFMs: Amazon Chronos \cite{ansari2024chronoslearninglanguagetime}, Lag-Llama \cite{rasul2024lagllamafoundationmodelsprobabilistic}, and TimesFM \cite{das2024decoderonlyfoundationmodeltimeseries} on four datasets: ERCOT, NN5 Daily, NN5 Weekly, and M3 Monthly \cite{gluonts_jmlr}. These models and datasets were chosen for zero-sot evaluation purposes (i.e. the models have never seen the benchmarking datasets). We compare and evaluate the performance of these models relative to statistical and gradient boosting models. 

This article is organized as follows: Section \ref{sec:tsfm} provides a concise overview of various FMs and the challenges they face in time series analysis, followed by the introduction of conformal prediction in Section \ref{sec:scp}. Section \ref{sec:experiments} details the experimental procedures. We review related work in Section \ref{sec:related_work}, and conclude the article with a discussion of our findings in Section \ref{sec:discussion}.

\section{Foundation models for time series forecasting}
\label{sec:tsfm}

Foundation models (FMs) are large-scale machine learning models trained on vast amounts of data, enabling them to generalize across multiple tasks with minimal or no fine-tuning \cite{bommasani2021opportunities}. 
These models have recently demonstrated remarkable success, particularly in natural language processing (NLP), computer vision (CV), and multimodal \cite{zhou2024comprehensive}. However, applying FMs to time series data presents unique challenges due to the inherent complexities of temporal structures:
\begin{itemize}
    \item Most of FMs are built upon the transformer architecture \cite{vaswani2017attention}. The \textbf{continuity} of time series data presents a fundamental challenge for transformer-based models, which are originally designed for discrete token-based inputs, such as those in natural language processing (NLP). Unlike NLP tasks, where predictions are made based on a finite and well-defined vocabulary of tokens, time series data is continuous, requiring models to capture subtle temporal dependencies and variations over an unbounded numerical space. This continuous nature increases the complexity of learning meaningful representations, making it more difficult for transformers to generalize effectively compared to tasks involving discrete sequences.
    \item In time series forecasting, \textbf{covariates} play a crucial role in improving predictive accuracy. These covariates can be static (e.g., demographic information), dynamic (e.g., weather conditions), or even additional time series that evolve alongside the primary series. However, most forecasting models struggle to effectively incorporate covariates, especially when the covariates themselves are time series that require simultaneous prediction. This scenario, known as multivariate time series forecasting, adds another layer of complexity. 
    \item  While NLP, CV, and multimodal benefit from an abundance of publicly available datasets \cite{deng2009imagenet, wang2018glue, lin2014microsoft}, time series forecasting faces a significant challenge due to the \textbf{limited availability of high-quality, diverse datasets}. This scarcity hampers the development and generalization of forecasting models, as they often rely on domain-specific or proprietary data \cite{lim2021time}.
    \item Most importantly, the \textbf{semantic aspect} of time series data is different than that of other data. In NLP, individual words carry inherent meaning, even without context, allowing models to leverage predefined semantic relationships. In contrast, time series data consists of ordered sequences where meaning emerges from the variability and structure of the data rather than from individual points. A single data point in a time series holds little significance on its own, as its interpretation depends entirely on the surrounding temporal context and underlying trends. This fundamental difference poses a unique challenge for time series models, requiring them to capture dependencies across time rather than relying on discrete semantic units.
\end{itemize}

Nevertheless, the development of foundation models for time series forecasting is currently a highly active field and is receiving significant attention across academia and industry. Several researches adapted the transformer architecture to create TSFMs, including GPT4TS \cite{zhou2023fitsallpowergeneraltime}, MOIRAI \cite{woo2024unifiedtraininguniversaltime}, TTMs \cite{ekambaram2024tinytimemixersttms}, Timer \cite{liu2024timergenerativepretrainedtransformers}, MOMENT \cite{goswami2024momentfamilyopentimeseries}, GTT \cite{feng2024curveshapematterstraining}, Time-MoE \cite{xioming2024timemoe}, Sundial \cite{liu2025sundialfamilyhighlycapable}. The landscape is rapidly evolving rapidly with new models emerging almost weekly or monthly. In this paper, we have selected three TSFMs for our experiments:

\begin{itemize}
    \item \textbf{Lag-Llama} is a decoder-only transformer model tailored for univariate time series forecasting that distinguishes itself by explicitly incorporating lagged values and calendar features into its input representation. Instead of solely relying on implicit temporal dependencies via self-attention, Lag-Llama engineers its input tokens to include recent observations (e.g., values from the previous day, week, or year) alongside time-specific indicators, thereby directly embedding periodic and seasonal patterns into the model. Lag-Llama uses language model LLaMA \cite{touvron2023llama} as the backbone. It employs causal self-attention mechanisms with normalization and rotary positional encoding to maintain temporal order, and it outputs a probability distribution (modeled as a Student-t distribution) for the next time step’s value using a dedicated distribution head. This design facilitates efficient, probabilistic forecasting in a lightweight framework, though its autoregressive, single-step prediction approach can accumulate error over long horizons.
    
    \item \textbf{Chronos} \cite{ansari2024chronoslearninglanguagetime} reinterprets time series forecasting as a language modeling problem by converting continuous series into discrete tokens using a quantization scheme that maps scaled numerical values into a finite vocabulary (typically 4096 tokens), thereby enabling the use of a transformer architecture derived from the T5 encoder–decoder framework \cite{raffel2020exploring}. Its architecture processes input sequences through an encoder that embeds the quantized tokens and a decoder that autoregressively generates future tokens. Chronos captures temporal dependencies via self-attention and positional encoding without explicitly incorporating exogenous temporal features. Training is performed with a standard cross-entropy loss over token sequences, and inference involves iterative token sampling that is later de-quantized to reconstruct continuous forecasts. Although this approach leverages massive pretraining on diverse time series data to yield strong zero-shot performance, the sequential generation of forecasts can be computationally intensive over long horizons, a drawback partially mitigated by optimizations in a recent version (Chronos-Bolt).

    \item \textbf{TimesFM} \cite{das2024decoderonlyfoundationmodeltimeseries} is a decoder-only transformer designed for long-horizon forecasting that innovates by segmenting time series data into patches—groups of consecutive time steps—that serve as the fundamental tokens for the model. By embedding these patches with positional information and processing them through a transformer with causal self-attention, TimesFM efficiently captures both short-term intra-patch patterns and long-term inter-patch dependencies, drastically reducing the effective sequence length and computational overhead compared to stepwise approaches. The model is trained using a regression objective (mean squared error) to predict entire blocks of future values in a single forward pass. As a result, TimesFM is fast, even for forecasting over extensive horizons.  Although its substantial parameter count (around 200 million) and extensive pretraining on diverse datasets afford it robust zero-shot performance across various domains, TimesFM’s lack of intrinsic probabilistic output necessitates auxiliary techniques for uncertainty estimation, positioning it as a highly efficient yet deterministically focused solution for long-term prediction tasks.

\end{itemize}




These models were selected because of the following resons:
\begin{itemize}
    \item None of these three models had already seen our benchmarking datasets (see Section \ref{subsec:datasets}) during their training phase. This allows us to evaluate their performance in a zero-shot prediction fashion.
    \item Chronos and TimesFM were selected due to their outstanding performance reports and frequent citations in related research to TSFMs \cite{ekambaram2024tinytimemixersttms, xioming2024timemoe, liu2025sundialfamilyhighlycapable} 
    \item Lag-Llama was chosen because of its size. It is a very small model compared to the other two (see Table \ref{tab:model_description}). We will show later in this paper the relation between the size of TSFMs and their performance. 
\end{itemize} 

Our primary objective is not to compare the performance of the TSFMs with each other but rather to assess the performance of TSFMs relative to traditional forecasting methods in a CP setting.

\section{Conformal prediction}
\label{sec:scp}

Conformal prediction \cite{shafer2007tutorialconformalprediction} is a robust statistical method for uncertainty quantification that can be applied to any forecasting model. CP uses a subset of past data, so-called calibration data,  to calculate a conformity measure, which assesses how well the predictive model performs on new data. This measure is then used to construct prediction intervals that adapt to the underlying data distribution. In time series modeling, CP provides a principled approach to generating prediction intervals with a specified confidence level over time.

There are several methods for constructing conformalized prediction intervals for time series forecasting, including Split Conformal Prediction (SCP) \cite[]{lei2017splitconformalprediction}, Adaptive Conformal Inference (ACI) \cite[]{gibbs2021adaptiveconformalinferencedistribution}, and Ensemble Batch Prediction Intervals (EnbPI) \cite[]{xu2021conformal}. These methods differ primarly in how they construct the calibration set to account for the temporal dependencies in sequential data when generating prediction intervals. In this work, we adopt Split Conformal Prediction (SCP) due to its simplicity, computational efficiency, and broad applicability. 


SCP \cite[]{lei2017splitconformalprediction} is a computationally efficient method to construct CP. It operates by splitting the available historical data into two disjoint subsets: (i) a training set to train the predictive model and (ii) a calibration set to estimate the uncertainty associated with the model’s predictions.

Given a user-defined \textit{miscoverage rate} $\alpha \in [0, 1]$ and  $N$ data points $(x_i, y_i) \in \mathbb{R}^d \times \mathbb{R}$, $i \in \{ 1,.., N \}$, SCP constructs the predictions intervals that cover the true values at least $((1-\alpha) \times 100)\%$ of the time over the prediction horizon $H$, using the following procedure:
\begin{itemize}
    \item \textit{Step 1 (splitting}): split $N$ available data points into two subsets: a training set $Tr = \{(x_i, y_i) | i \in  \{ 1,.., N_s\}\} $ and a calibration set $Cal = \{(x_i, y_i) | i \in  \{ N_s+1,..,N\}\}$, with $1 < N_s < N$.
    
    \item \textit{Step 2 (training)}: train a regressor $\hat{f}$ using the training set $Tr$.
    
    \item \textit{Step 3 (\textit{conformity score})}: use the fitted regressor $\hat{f}$ to make predictions on the calibration set $Cal$, then calculate the conformity between the true values and the predicted ones: $S_{Cal} = \{S_{i}\}_{i \in Cal} $ , with $S_{i} = |\hat{f}(x_{i}) - y_{i}|$.
    
    \item \textit{Step 4 (uncertainty threshold)}: the uncertainty threshold $\hat{q}$ is defined as the $q_{1-\hat{\alpha}}$ quantile of the conformity scores:
    \begin{equation}
        \hat{q} = q_{1-\hat{\alpha}}(S_{Cal})
        \label{eq:uncertainty_threshold}
    \end{equation}
    with $1-\hat{\alpha} = \frac{ \left\lceil(\left| Cal \right|+1)(1-\alpha)
    \right\rceil}{\left| Cal \right|}$, $\left| Cal \right|$ is the cardinality of $Cal$ and $\lceil . \rceil$ denotes the ceiling function.
    
    \item \textit{Step 5 (prediction interval)}: 
    the prediction interval $I(x_{N+k})$ at time step $N+k$ on the test set are calculated as: 
    \begin{equation}
        \label{eq:intervals}
        I(x_{N+k}) = [\hat{f}(x_{N+k})-\hat{q}, \hat{f}(x_{N+k})+\hat{q}], \quad \text{for} \quad k \in \{1,..,H \}.
    \end{equation}
\end{itemize}

Although other methods, such as Conformalized Quantile Regression (CQR) \cite{romano2019conformalizedquantileregression} might improve the estimation of the prediction interval, we use SCP in this work for its simplicity.

\textbf{Global vs local quantiles}: when a dataset contains several time series, two options could be considered:
\begin{itemize}
    \item $\hat{q}$ is computed across all time series. This approach is dubbed as ``global quantile'' in this paper.
    \item $\hat{q}^j$ is computed for each time series $j$. This approach is dubbed as ``local quantile'' in this paper.
\end{itemize}

Formally, suppose that we have $M$ time series in the dataset. The train/calibration splitting is applied to each series, we then have $M$ sets $\{Tr^{j}, Cal^{j}\}$, for $j \in \{1,.., M\}$. 
The global quantile is defined as:
$$ \hat{q}_{global} = q_{1-\hat{\alpha}}(\cup S_{Cal^{j}})|_{j \in \{1,.., M\}}$$
while one local quantile is calculated for each time series:
$$ \hat{q}^j = q_{1-\hat{\alpha}}(S_{Cal^{j}}), \quad \text{for} \quad j \in \{1,..,M \} $$

In this article, we report local quantile. The global quantile results can be found in Appendix \ref{apd:global_vs_local}.\\

\section{Experiments and Results}
\label{sec:experiments}

\subsection{Datasets}
\label{subsec:datasets}

In this study, we used the ERCOT, NN5 Daily, NN5 Weekly, and M3 Monthly as the benchmarking datasets. These datasets, spanning different temporal granularities, were selected to ensure that the results accurately reflect real-world applications. To evaluate the impact of the forecasting horizon, for each dataset, we conducted experiments across three horizons: Short (S), Medium (M), and Long (L). The details of the datasets and the corresponding horizons are provided in Table \ref{tab:datasets_description}.

\begin{table}[h!]
    \centering
    \renewcommand{\arraystretch}{1.5}
    
    \begin{tabular}{lcccc}
        \toprule
        \textbf{Datasets} & \textbf{Num of series} & \textbf{Series length} & \textbf{Frequency} & \textbf{Forecasting horizon (S, M, L)} \\
        \midrule
        ERCOT      & 1    & 17520 & H (hourly) & 24, 72, 168 \\
        NN5 Daily  & 100  & 791      & D (daily) & 7, 21, 35 \\
        NN5 Weekly & 100  & 105       & W (weekly) & 4, 8, 12  \\
        M3 Monthly & 1428 & 66       & M (monthly) & 3, 6, 9 \\
        \bottomrule
    \end{tabular}
    \vspace{0.1cm}
    \caption{Datasets description}
    \label{tab:datasets_description}
\end{table}

\subsection{Estimators}
\label{models}
We evaluated three TSFM candidates mentioned above againt various TS forecasting methods, including the Naive method, Seasonal Naive method, LightGBM (LGBM) \cite{ke2017lightgbm} and a light version of StatisticalEnsemble \cite{scum} \cite{Nixtla2023}. For Chronos and TimesFM, we tested both the original versions, dubbed as Chronos and TimesFM, and the more recent ones, dubbed as Chronos Bolt and TimesFM2, respectively. We show later in this paper that during just a short period, Chronos and TimesFM have improved their performance significantly with the updates. LGBM was selected as an representative of the Gradient Boosted Decision Trees (GBDT) family, of which XGBoost \cite{chen2016xgboost} or CatBoost \cite{prokhorenkova2019catboostunbiasedboostingcategorical} are wildly popular models. StatisticalEnsemble  is an ensemble of statistical models, consisting of the mean of forecasts from AutoARIMA, AutoETS, AutoCES and DynamicOptimizedTheta. These models were chosen for their good performances in the M4 competition \cite{MAKRIDAKIS202054}.  However, due to its high computational complexity, AutoARIMA is not used in this study. 

The detailed configuration is as follows:
\begin{itemize}
    \item The \textbf{Naive} method forecasts each point in the prediction horizon by utilizing the value of the most recently observed point. All available data points were used for calibration, with the exception of the first one, which serves as context.
    \item The \textbf{Seasonal Naive} method predicts each point in the prediction horizon by utilizing the values of the last known season. The seasonality parameter is adjusted to match the prediction horizon. All available data points were used for calibration, with the exception of the first ones, which serves as context.
    \item \textbf{StatisticalEnsemble\_light} consists of three models (AutoETS, AutoCES, and DynamicOptimizedTheta), all configured similarly. The seasonality parameter is set according to the frequency of each dataset: 24 for hourly, 7 for daily, 4 for weekly, and 12 for monthly data. The data were split 80\% for training and 20\% for calibration.
    \item For \textbf{LGBM}, we tested three different train-calibration ratios: 80\%-20\%, 50\%-50\%, and 20\%-80\%. As input features, lag variables were created based on the time series frequencies and prediction horizons. Additionally, datetime features were included. Missing values resulting from lag creation were dropped. We used the defaut values for the hyperparameters. 
    \item For \textbf{TSFMs}, we employed the setup described in Table \ref{tab:model_description}. All available data points were divided into context and calibration points. The context length was chosen based on data availability and set as a multiple of 32. When sufficient data points were available, the context length was set to 512. If fewer data points were available, it was reduced to as low as 32 in the worst case. A dataset was considered to have enough data points when the context length was 8 to 10 times the prediction length, with at least 60\% of the points allocated for calibration. 
\end{itemize}

\begin{table}[h!]
    \label{tab:choice_context}
    \centering
    \renewcommand{\arraystretch}{1.5} 
    \begin{adjustbox}{max width=\textwidth}
    \begin{tabular}{lcccc} 
    \toprule
    \textbf{Model} & \textbf{\# parameters} & \textbf{\# training data} & \textbf{Loss function} & \textbf{Output} \\ 
    \midrule
    Lag-Llama         & 2.5M  & 352M & Neg log-likelihood    & Student's t-distribution \\
    Chronos (chronos-t5-small)   & 46M & 84B  & Cross entropy        & Categorical distribution               \\
    Chronos-bolt (chronos-bolt-small)   & 48M & \textit{unknown}  & Cross entropy        & Categorical distribution               \\
    TimesFM (timesfm-1.0-200m)           & 200M & $\mathcal{O}(100B)$  & Mean of MSE's                   & Point forecast         \\
    TimesFM 2.0 (timesfm-2.0-500m)           & 500M & \textit{unknown} & Mean of MSE's                   & Point forecast         \\
    \bottomrule
    \end{tabular}
    \end{adjustbox}
    \vspace{0.1cm}
    \caption{FMs setup description. }
    \label{tab:model_description}
\end{table}

\begin{figure}[h!]
    \centering
    \includegraphics[width=\textwidth]{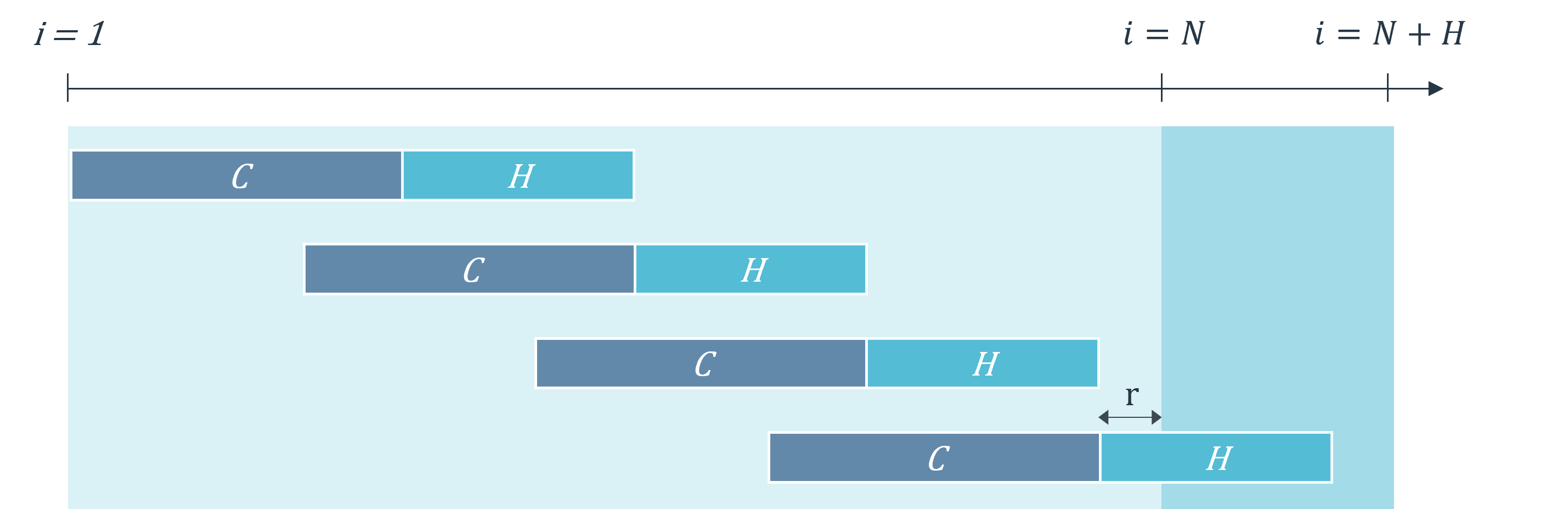}
    \caption{Rolling window. Let $C$ the context size and $H$ the prediction length. We use the first $C$ available data points as the context to predict $H$ points $\{\hat{y}_{C+1}, ... \hat{y}_{C+H}\}$. The window is then shifted forward by $H$ steps to predict the next set of points $\{\hat{y}_{C+H+1}, ... \hat{y}_{C+2H}\}$. This process continues until only $r$ data points remain, where $0\le r < H$. In the final step,  $H$ points are forecasted with context $C$, but only the first $r$ points are retained for the calculation of the conformity score.}
    \label{fig:rolling_window}
\end{figure}

We used the rolling window strategy shown in Figure \ref{fig:rolling_window} to calibrate our models.

The only methods that do not use a rolling window approach for forecasting are \textbf{LGBM} and \textbf{StatisticalEnsemble\_light}. These methods divide the $N$ data points into a training set and a calibration set, with a given percentage of data allocated to each set. 


\subsection{Metrics}
\label{subsec:metrics}



We evaluated the models using three metrics: mean coverage rate $MCR$, mean scaled interval width $MSIW$, and mean absolute scaled error $MASE$. The details are as follows:

\subsubsection{Mean Coverage Rate (MCR)}
\label{subsec:covarage_rate}

The \textbf{Coverage Rate} is the proportion of times that the true value falls within the prediction interval. For each series $j$ in the test set, it is defined as follows:

\begin{equation}
    CR^j =\frac{1}{H} \sum_{i=1}^{H} \mathds{1}\{ y^j_{i} \in I^j_{i} \}
    \label{eq:coverage_rate_j}
\end{equation}

where $H$ is the prediction horizon, $y^j_{i}$ and $I^j_{i}$ are the true value and the prediction interval, respectively, at time $i$ of series $j$.  $\mathds{1}\{ \cdot \}$ is the indicator function that is 1 if the condition inside $\{ \cdot \}$ is true and 0 otherwise.

We then average the coverage over the test set to get the mean coverage rate:

\begin{equation}
    MCR = \frac{1}{M_{test}} \sum_{j=1}^{M_{test}} CR^j
    \label{eq:coverage_rate}
\end{equation}
with $M_{test}$ the number of series in the test set. 

\subsubsection{Mean Scaled Interval Width (MSIW)}
\label{subsec:interval_width}

The \textbf{Interval Width} measures the width of the prediction interval. Formally, for each series $j$ in the test set, the interval width is defined as follows:

\begin{equation}
    IW^j = \frac{1}{H} \sum_{i=1}^{H} \left( I^{j}_{upper, i} - I^{j}_{lower, i} \right)
    \label{eq:interval_width}
\end{equation}

where $I^{j}_{upper, i}$ and $I^{j}_{lower, i}$ are the upper and lower bounds,  respectively, of the prediction interval at time step $i$ of series $j$.

\begin{figure}[h!]
    \centering
    \includegraphics[width=0.8\textwidth]{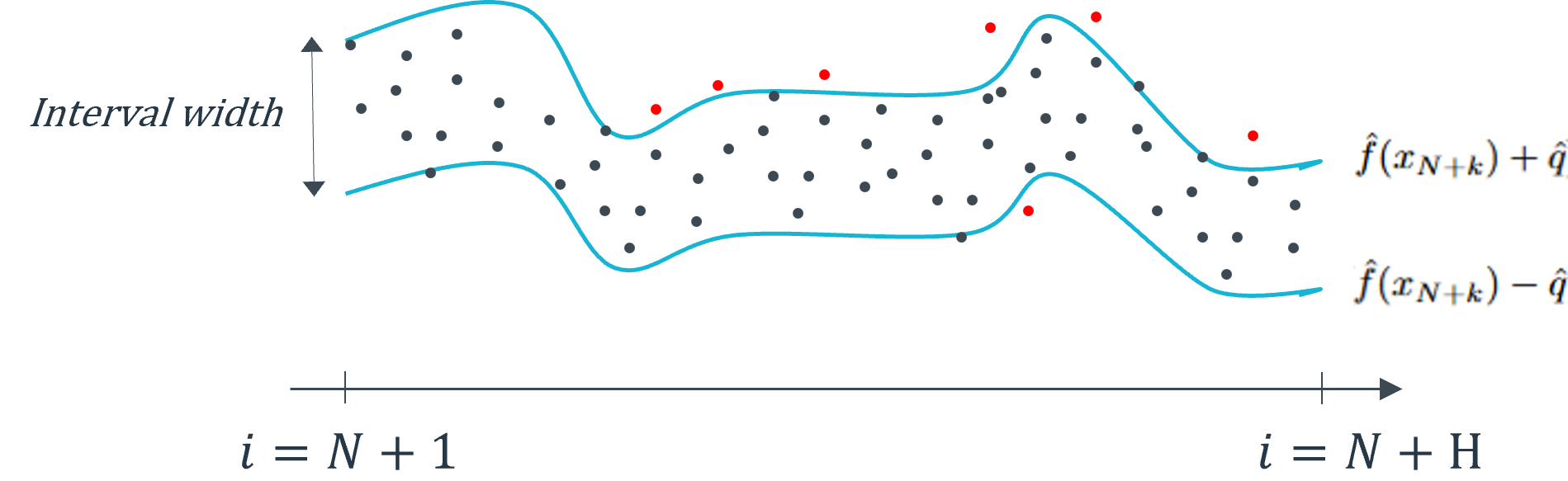}
    \caption{Illustration of prediction interval. The light blue lines represent the upper and lower bounds of the prediction interval, within which future data points are expected to fall with a specified probability. The scattered dots represent individual data points, with black dots indicating those that fall within the prediction interval and red dots indicating those that fall outside. Coverage refers to the fraction of data points that lie within the interval, while interval width is the distance between the upper and lower bounds, representing how wide the prediction interval are on average across the domain.}
    \label{fig:prediction_interval_explaination}
\end{figure}

An illustration of the interval width, as well as the coverage rate is shown in Figure \ref{fig:prediction_interval_explaination}.

We aggregate the interval widths over the test set to get the \textbf{Mean Scaled Interval Width} as follows:

\begin{equation}
    MSIW = \frac{1}{M_{test}}\sum^{M_{test}}_{j=1}\frac{IW^j}{IW^j_{naive}}
    \label{eq:isw}
\end{equation}
where $IW^j_{naive}$ the width of the prediction interval produced by the Naive model.


\subsubsection{Mean Absolute Scaled Error (MASE)}
\label{subsec:mase}

The \textbf{Mean Absolute Error} of a model $f$ on series $j$ is defined as follows:

\begin{equation}
    MAE_j = \frac{1}{H}\sum_{i=1}^{H} |f(x^j_{i}) - y^j_{i}|.  
    \label{mse}
\end{equation}

Similarly to the $MISW$, we aggregated the mean absolute error over the test set to get the \textbf{Mean Absolute Scaled Error} as follows:

\begin{equation}
    MASE =  \frac{\sum^{M_{test}}_{j=1}MAE_j}{\sum^{M_{test}}_{j=1}MAE_{naive, j}}
    \label{eq:isw}
\end{equation}
with $mae_{naive, j}$ the mean absolute error of the naive model on series $j$.

\subsection{Results on the ERCOT dataset}
\label{subset:ercot}

\textbf{Description}: 
In this first experiment, we evaluated the performance of the benchmarking forecasting models on the ERCOT dataset. To simulate a situation where data is not available in large amounts, we used only a subset from the years 2018 and 2019 as historical data instead of the whole dataset. As the data is at an hourly granularity, the resulting set contains a total of 17520 data points. From this historical data, 20 windows were sampled. Each model was applied and tested on each window to ensure robust results. The final results are the aggregation over these 20 windows.

We set up two different scenarios, each with a different amount of available data, in order to observe how the different methods studied behave:

\begin{itemize}
    \item Scenario 1:
    Each window is composed of 365 days (8760 data points) plus the prediction horizon. The 365 days correspond to the available data for the training and calibration phases. The prediction horizons are 1, 3, and 7 days (24, 72, and 168 data points, respectively). The final result is the aggregation over these 20 windows.
    \begin{figure}[h!]
        \centering
        \includegraphics[width=\textwidth]{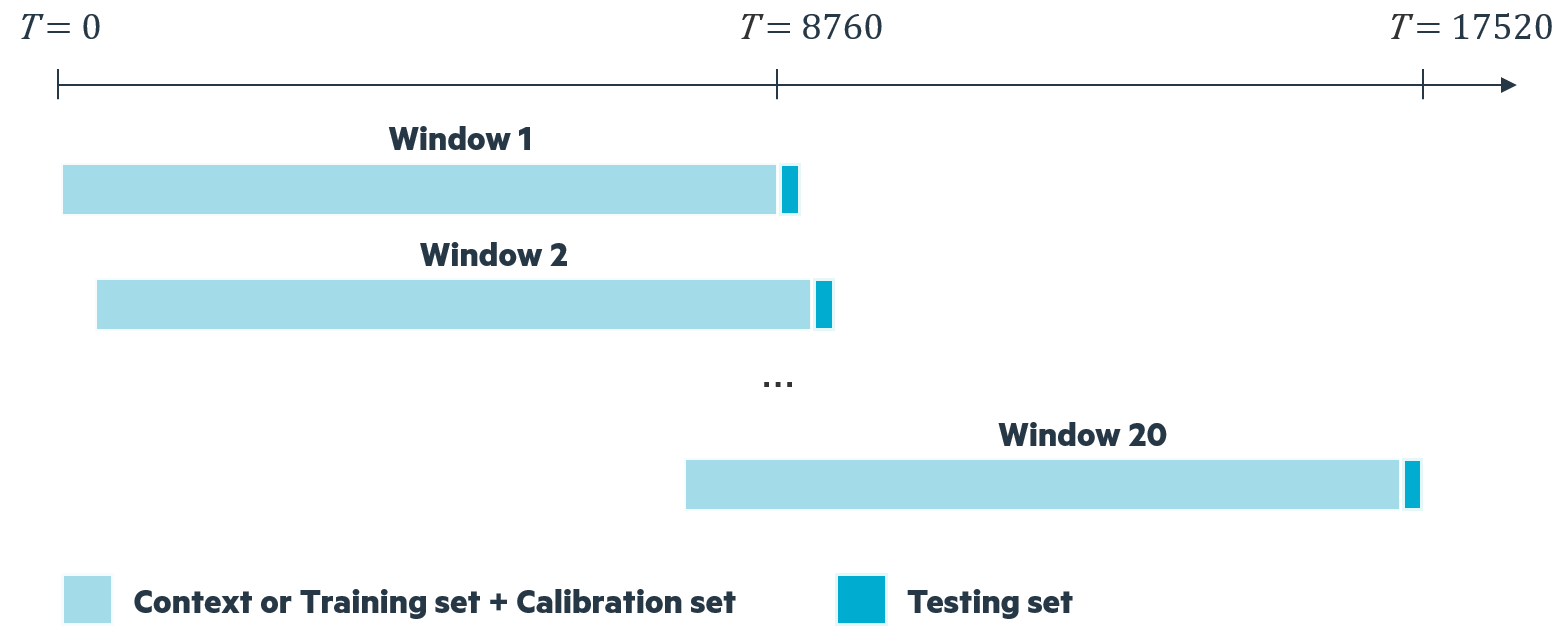}
        \caption{Window sampling - Scenario 1}
    \end{figure}
    \item 
    Scenario 2: 
    Each window is composed of 93 days (2232 data points) plus the prediction horizon. The 93 days correspond to the available data for the training and calibration phases. The prediction horizons are 1, 3, and 7 days (24, 72, and 168 data points, respectively). The aggregated results of these 20 windows will be presented. 

\end{itemize}

Approximately two years of ERCOT data were selected as the test set to ensure that it spans the chosen period, capturing seasonal variations with minimal overlap between windows. We assume that 20 windows are sufficient to provide a reliable estimate of the models' performance on the ERCOT dataset.

\begin{table}[h!]
\centering
\begin{adjustbox}{max width=\textwidth}
\begin{tabular}{lcccc}
\toprule
\multirow{2}{*}{\textbf{Model}} & \multicolumn{2}{c}{\textbf{ERCOT 8760 points}} & \multicolumn{2}{c}{\textbf{ERCOT 2232 points}} \\
\cmidrule(lr){2-3} \cmidrule(lr){4-5}
                                & \textbf{Points for context/fit} & \textbf{Points for calibration} 
                                & \textbf{Points for context/fit} & \textbf{Points for calibration} \\
\midrule
Naive                     & 1                     & 8759                & 1                     & 2231                \\
SeasonalNaive             & h                     & 8760-h                & h                     & 2232-h                \\
StatisticalEnsemble.light & 7008                  & 1752                & 1785                  & 447                 \\
LGBM\_20\_80              & 0.2x(8760-h-168)      & 0.8x(8760-h-168)    & 0.2x(2232-h-168)      & 0.8x(2232-h-168)    \\
LGBM\_50\_50              & 0.5x(8760-h-168)      & 0.5x(8760-h-168)    & 0.5x(2232-h-168)      & 0.5x(2232-h-168)    \\
LGBM\_80\_20              & 0.8x(8760-h-168)      & 0.2x(8760-h-168)    & 0.8x(2232-h-168)      & 0.2x(2232-h-168)    \\
Lag-Llama                 & 512                   & 8248                & 512                   & 1720                \\
Chronos                   & 512                   & 8248                & 512                   & 1720                \\
ChronosBolt               & 512                   & 8248                & 512                   & 1720                \\
TimesFM                   & 512                   & 8248                & 512                   & 1720                \\
TimesFM2                  & 512                   & 8248                & 512                   & 1720                \\
\bottomrule
\end{tabular}
\end{adjustbox}
\vspace{0.1cm}
\caption{Model configuration for ERCOT.}
\label{tab:example}
\end{table}

\begin{figure}[h!] 
\includegraphics[width=\textwidth]{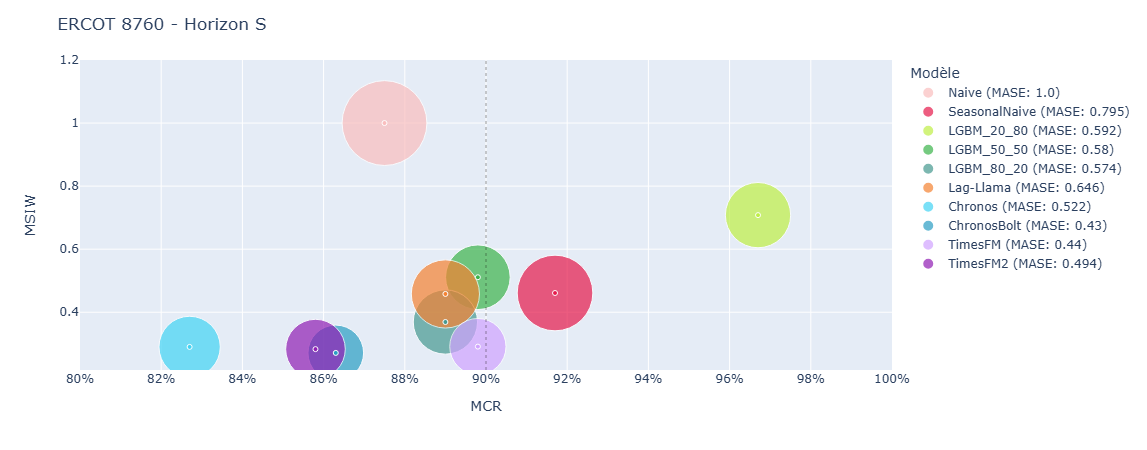} \includegraphics[width=\textwidth]{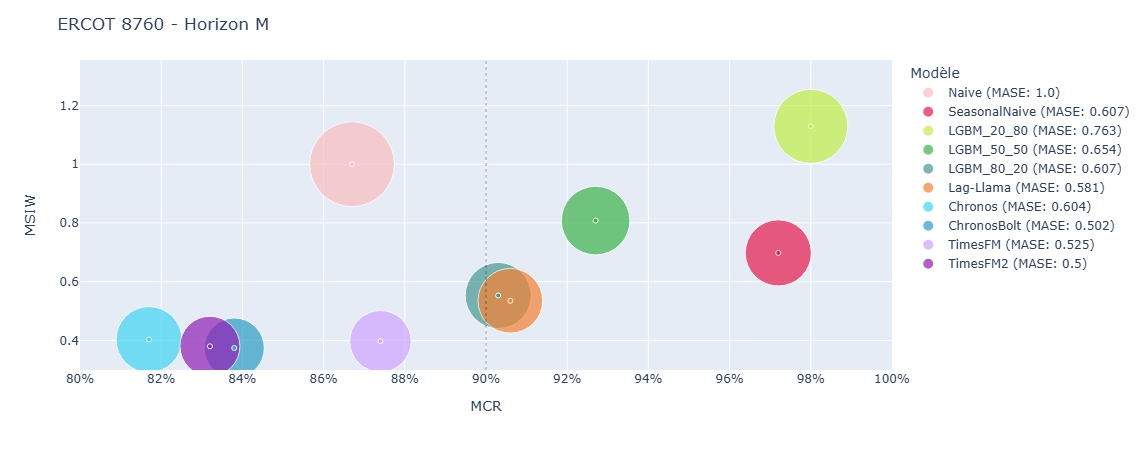} \includegraphics[width=\textwidth]{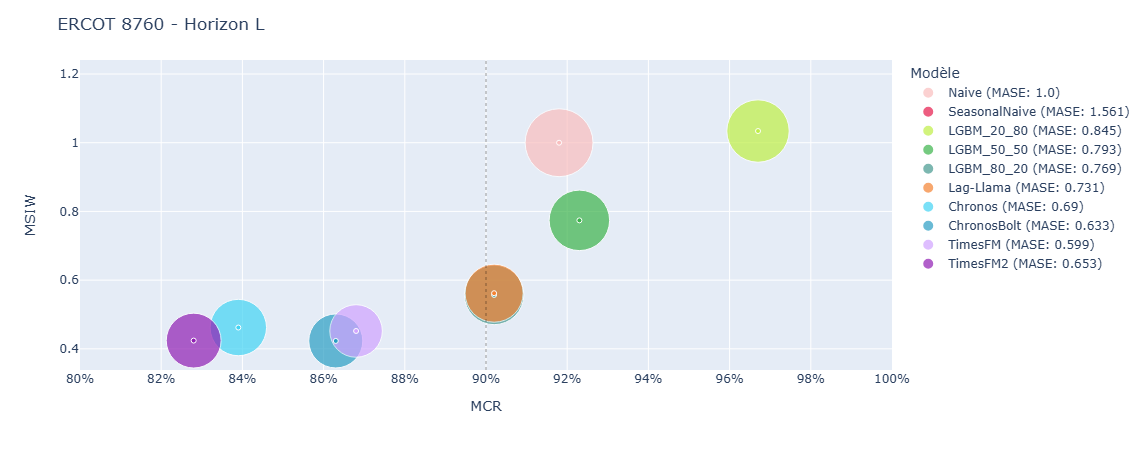} \caption{Split conformal prediction results on ERCOT 8760. Better models are those appear to the right-hand side of the 90\% vertical dashed line (indicating that they achieve the target coverage rate) or very close to it, at the bottom of the figure (indicating lower MSIW), and with smaller radius (indicating smaller MASE).} \label{fig:res1.1} \end{figure}

\begin{figure}[h!] \includegraphics[width=\textwidth]{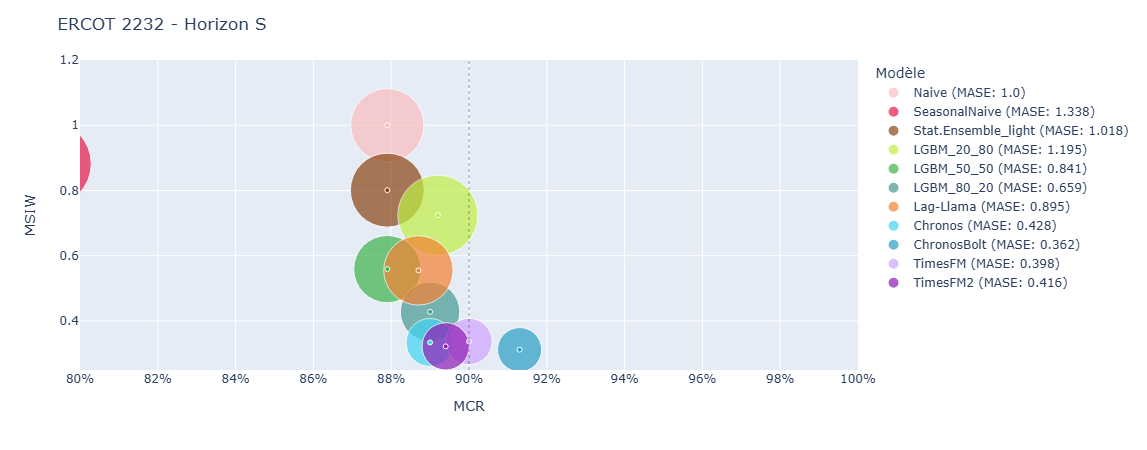} \includegraphics[width=\textwidth]{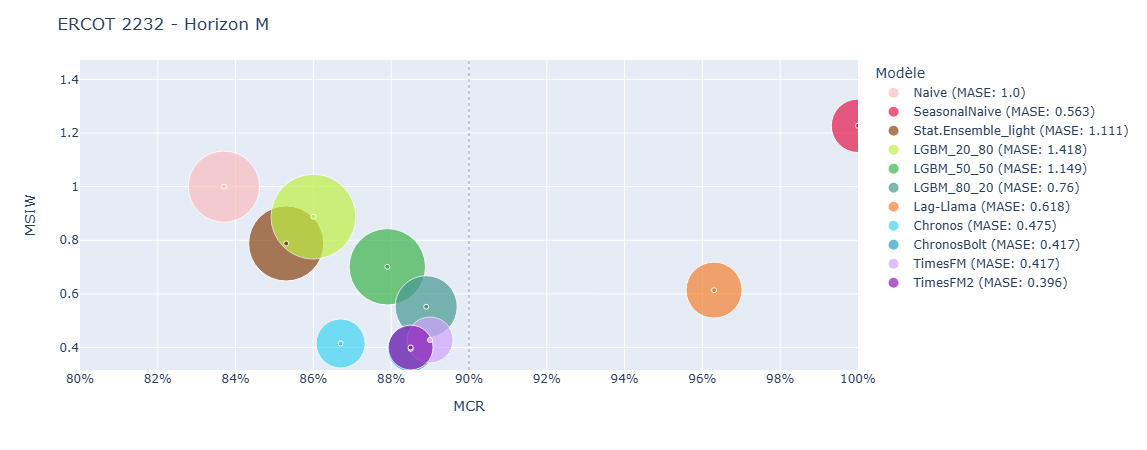} \includegraphics[width=\textwidth]{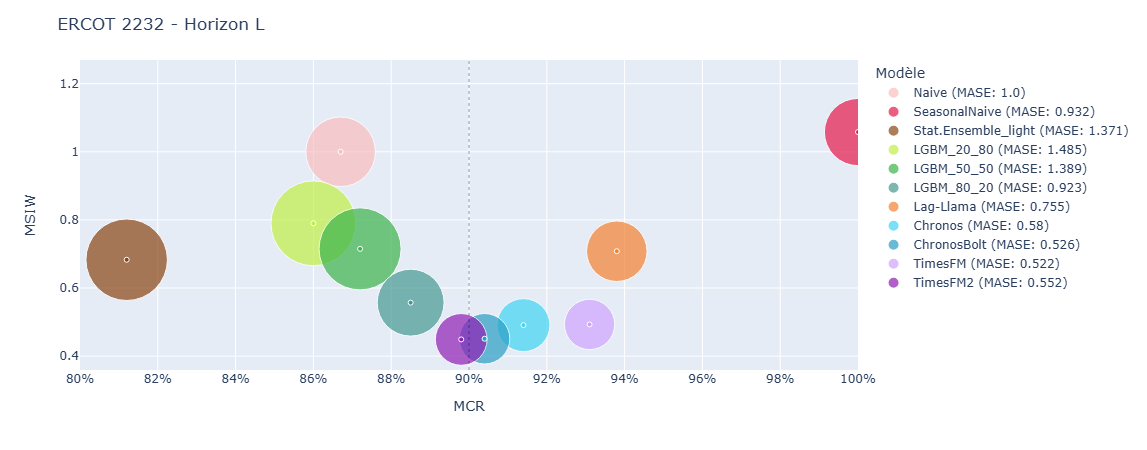} \caption{Split conformal prediction on ERCOT 2232} \label{fig:res1.2} \end{figure}

\textbf{Results}:
\label{sec:results}
Figures \ref{fig:res1.1} and \ref{fig:res1.2} show the performance of the benchmarking models in terms of MCR, MSIW, and MASE. The miscoverage rate $\alpha$ was set to 0.1 for a target coverage of 90\%. 
Several models did not reach that target (e.g. ChronosBolt, TimesFM, and TimesFM2 in the ERCOT-8760-M, ERCOT-8760-L settings). Those who reached the target gave in general large interval width and large MASE (SeasonalNaive, LGBM\_20\_80 in all the ERCOT-8760 settings, Lag-Llama and SeasonalNaive in the ERCOT-2232-M and ERCOT-2232-L settings). We argue that the coverage rate was not met because the data do not satisfy the exchangeability criterion of SCP \cite{lei2017splitconformalprediction}.

There was a positive correlation between the MASE (the size of each circle) and the MSIW (y-axis). However, this relationship didn't follow a linear pattern. In terms of MASE, ChronosBolt, TimesFM, TimesFM2, and Chronos stood out in all settings. For example, in the ERCOT-2232-S setting, ChronosBolt achieved the best MASE (0.362) and MSIW (0.312), why respecting the target coverage rate (91.3\%). LGBM\_80\_20 performed well and was comparable to ChronosBolt, TimesFM, TimesFM2. However, LGBM\_50\_50 and LGBM\_20\_80 yielded poor MASE in the ERCOT-2232-S setting (0.841 and 1.195, respectively), likely due to the relatively small amount of training data. These models performed significantly better in the 8760-point setup. Lag-Llama had slightly lower performance but remained stable, unlike SeasonalNaive. StatisticalEnsemble\_light performed poorly and was implemented only in the 2232-point setup due to time constraints.

In terms of coverage, TimesFM2, Chronos Bolt, and Chronos exhibited degraded coverage in the 8760-points setups. TimesFM gave better MSIW than the mentioned three. The other methods remained consistently close to the target coverage, mainly because the gave wide prediction intervals.


The complete results over different horizons are shown in Table \ref{ercot_results} in the Appendix. In general, TSFMs, especially Chronos, ChronosBolt, TimesFM, and TimesFM2, gave more meaningful predictions. They consistently achieved better MASE and MSIW. Althouh their coverage did not always meet the target, given that the prediction intervals of the other models were too wide, we are in favor of TSFMs for the overall prediction.

The performance of all the models decreased over longer horizons, as expected. This phenomenon was more pronounced for LGBM-based models in the 2232-point setups. We argue that LGBM suffers more from the lack of data. As the forecasting horizon increases, the creation of lag features results in the removal of missing values, leading to less data for LGBM. This negatively impacts performance, especially when the volume of input data is low.

\begin{figure}[htbp] 
    \centering 
    \includegraphics[width=\linewidth]{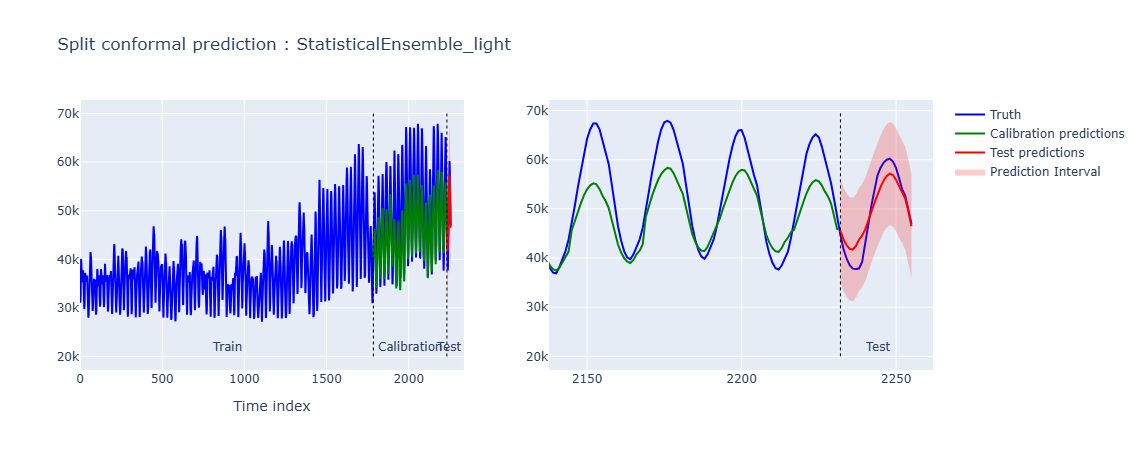} 
    \includegraphics[width=\linewidth]{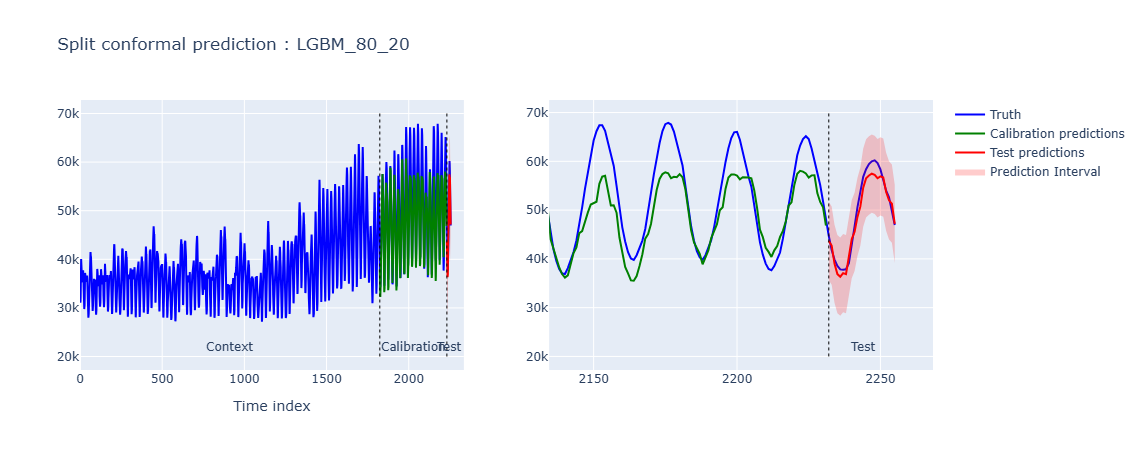} 
    \includegraphics[width=\linewidth]{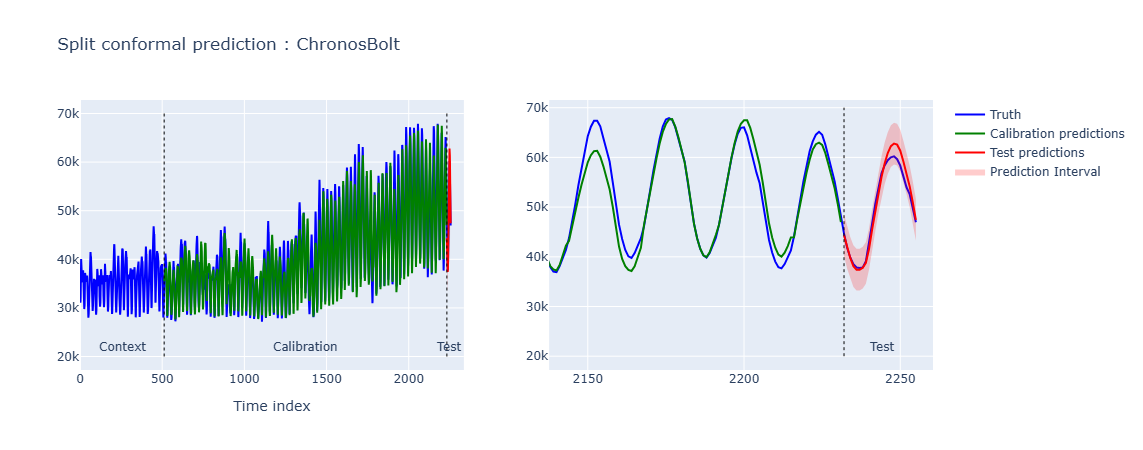} \includegraphics[width=\linewidth]{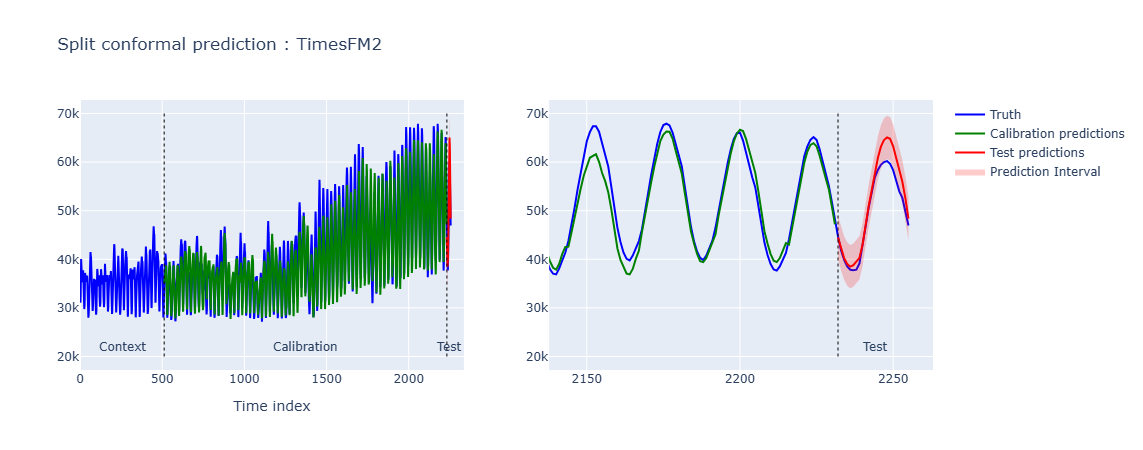} 
    \caption{Split conformal prediction on 2232 points of ERCOT - horizon S - with StatisticalEnsemble\_light, LGBM, Chronos Bolt, and TimesFM 2.} 
    \label{fig:res1} 
\end{figure}

Some examples of the predicted series are shown in Fig \ref{fig:res1}. All the models were able to capture the seasonality present in the data. However, StatisticalEnsemble\_light and LGBM\_80\_20 gave wider prediction intervals than Chronos Bolt and TimesFM2.

\clearpage

\subsection{Results on the NN5 Daily, NN5 Weekly, and M3 Monthly datasets}
\label{subsec:results_nn5_m3}
\textbf{Description}:
The first experiment evaluated the performance of our models on an univariate time series dataset. Subsequently, we aimed to compare their effectiveness on multivariate time series notably because, in the multivariate context, the number of available data points is determined not only by the number of available time steps but also by the number of available series.
Hence, we conducted a second experiment, on the NN5 Daily, NN5 Weekly, and M3 Monthly datasets. The models were configurated in the same manner as in the ERCOT experiment, with the details shown in Table \ref{tab:model_config_nn5_m3}.

\begin{table}[t]
\centering
\renewcommand{\arraystretch}{1.2} 
\begin{adjustbox}{max width=\textwidth}
\begin{tabular}{lc cc cc c}
\toprule
\multirow{3}{*}{\textbf{Model}} & \multicolumn{2}{c}{\textbf{NN5 Daily}} & \multicolumn{2}{c}{\textbf{NN5 Weekly}} & \multicolumn{2}{c}{\textbf{M3 Monthly}} \\
\cmidrule(lr){2-3} \cmidrule(lr){4-5} \cmidrule(lr){6-7}
& \textbf{Points for} & \textbf{Points for} & \textbf{Points for} & \textbf{Points for} & \textbf{Points for} & \textbf{Points for} \\
& \textbf{context/fit} & \textbf{calibration} & \textbf{context/fit} & \textbf{calibration} & \textbf{context/fit} & \textbf{calibration} \\
\midrule
Naive              & 1 & 791-1-h & 1 & 105-1-h & 1 & 66-1-h \\
SeasonalNaive      & h & 791-2xh & h & 105-2xh & h & 66-2xh \\
Stat.Ensemble\_light               & 0.8x(791-h) & 0.2x(791-h) & 0.8x(105-h) & 0.2x(105-h) & 0.8x(791-h) & 0.2x(66-h) \\
LGBM\_20\_80              & 0.2x(791-h-(h+28))x100 & 0.8x(791-h-(h+28)) & 0.2x(105-h-(h+8))x100 & 0.8x(105-h-(h+8)) & 0.2x(66-h-(h+6))x1428 & 0.8x(66-h-(h+6)) \\
LGBM\_50\_50              & 0.5x(791-h-(h+28))x100 & 0.5x(791-h-(h+28)) & 0.5x(105-h-(h+8))x100 & 0.5x(105-h-(h+8)) & 0.5x(66-h-(h+6))x1428 & 0.5x(66-h-(h+6)) \\
LGBM\_80\_20              & 0.8x(791-h-(h+28))x100 & 0.2x(791-h-(h+28)) & 0.8x(105-h-(h+8))x100 & 0.2x(105-h-(h+8)) & 0.8x(66-h-(h+6))x1428 & 0.2x(66-h-(h+6)) \\
Lag-Llama          & 128 & 791-128-h & 64 & 105-64-h & 32 & 66-32-h \\
Chronos    & 128 & 791-128-h & 64 & 105-64-h & 32 & 66-32-h \\
ChronosBolt    & 128 & 791-128-h & 64 & 105-64-h & 32 & 66-32-h \\
TimesFM            & 128 & 791-128-h & 64 & 105-64-h & 32 & 66-32-h \\
TimesFM2            & 128 & 791-128-h & 64 & 105-64-h & 32 & 66-32-h \\
\bottomrule
\end{tabular}
\end{adjustbox}
\vspace{0.1cm}
\caption{Model configuration for NN5 daily, NN5 Weekly, and M3 Monthly.}
\label{tab:model_config_nn5_m3}
\end{table}

\textbf{Results}: 
In this section, we present the benchmarking results on the three multivariate datasets introduced above. The results were aggregated over all time series in each dataset, rather than over 20 sample windows as in the ERCOT experiment described in the previous section.

\begin{figure}[h!]
    \includegraphics[width=\textwidth]{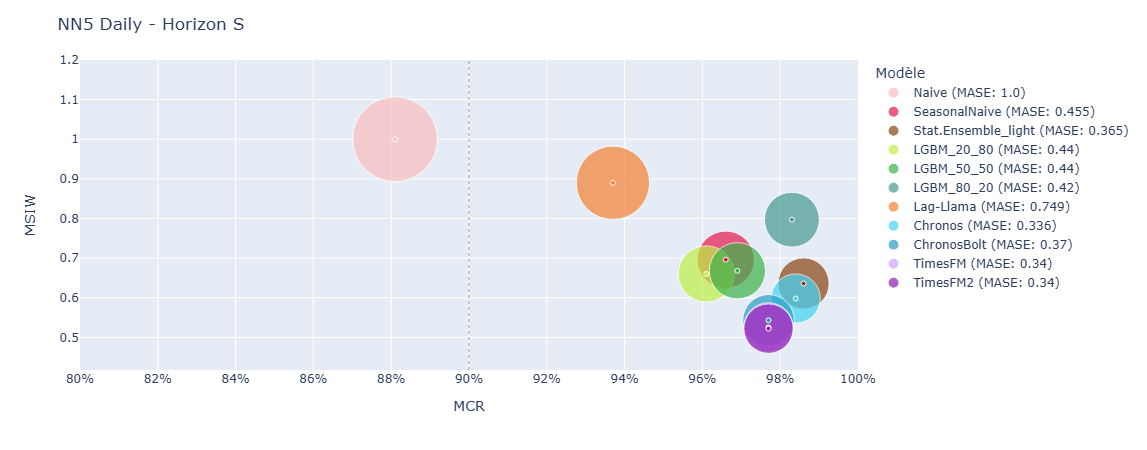}
    \includegraphics[width=\textwidth]{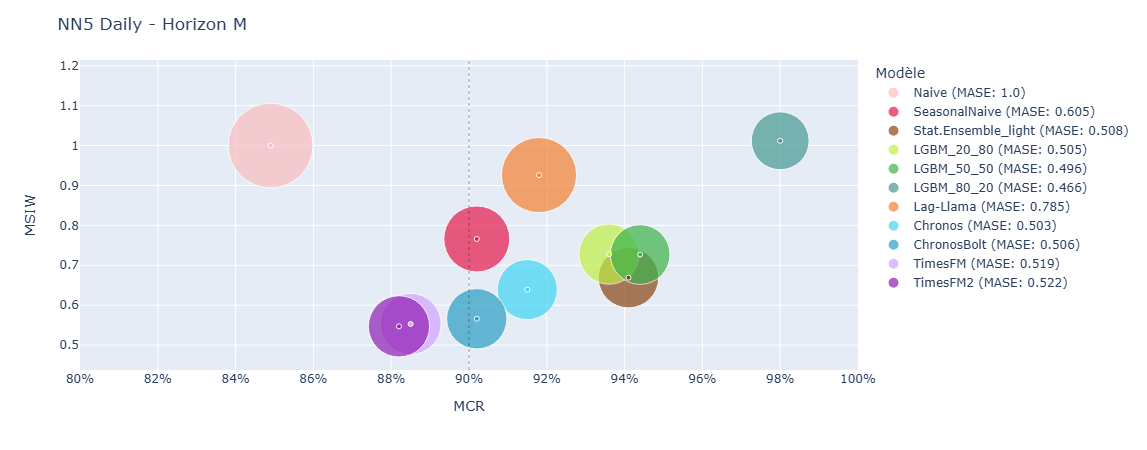}
    \includegraphics[width=\textwidth]{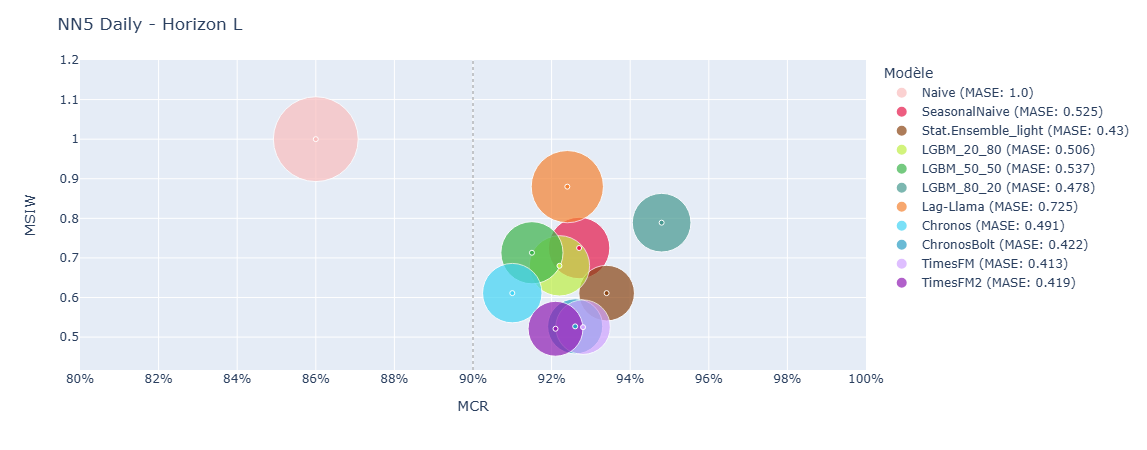}
    \caption{Split conformal prediction on NN5 Daily}
    \label{fig:nn5d_result}
\end{figure}

\begin{figure}[h!]
    \includegraphics[width=\textwidth]{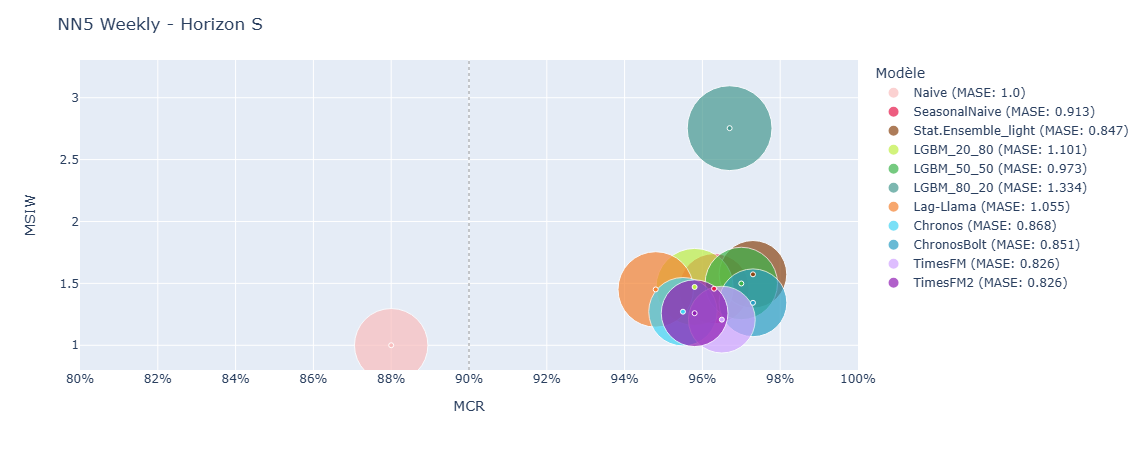}
    \includegraphics[width=\textwidth]{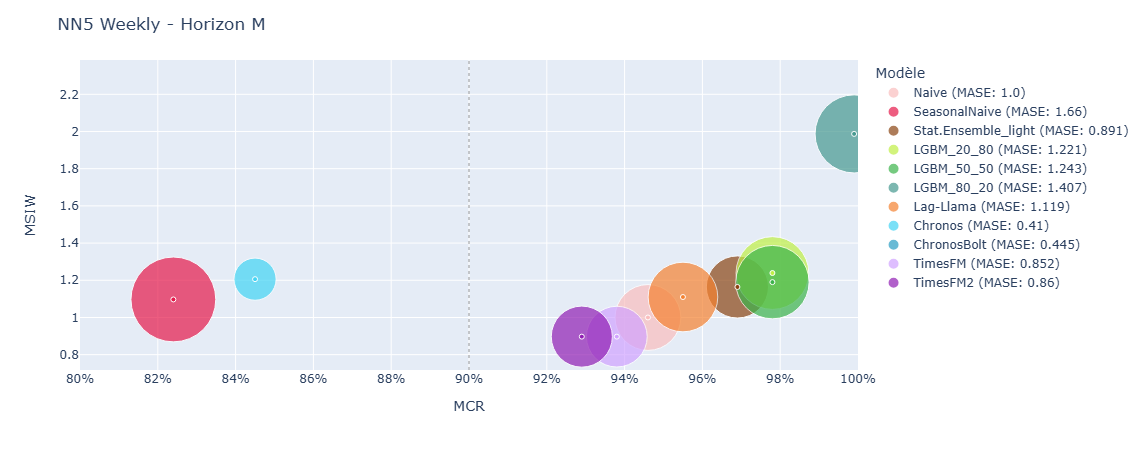}
    \includegraphics[width=\textwidth]{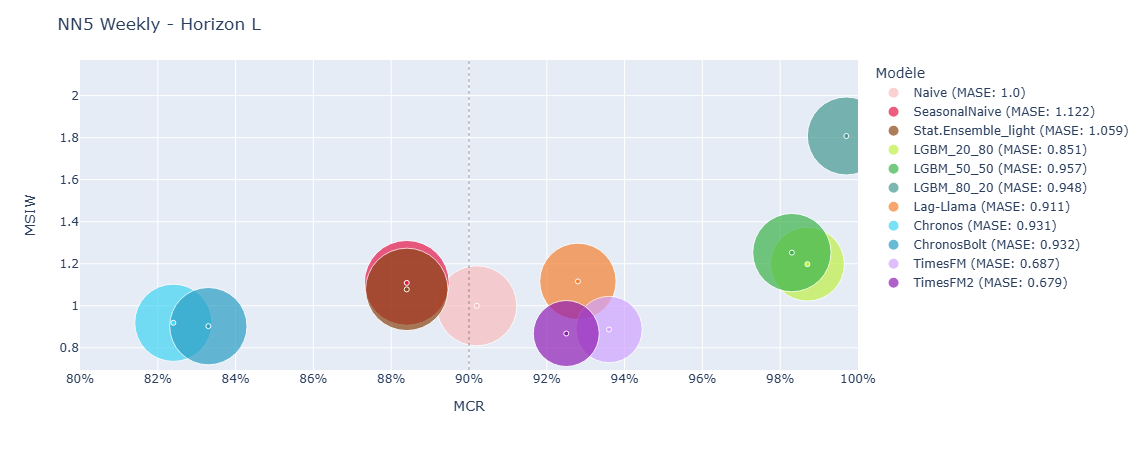}
    \caption{Split conformal prediction on NN5 Weekly}
    \label{fig:nn5w_result}
\end{figure}

\begin{figure}[h!]
    \includegraphics[width=\textwidth]{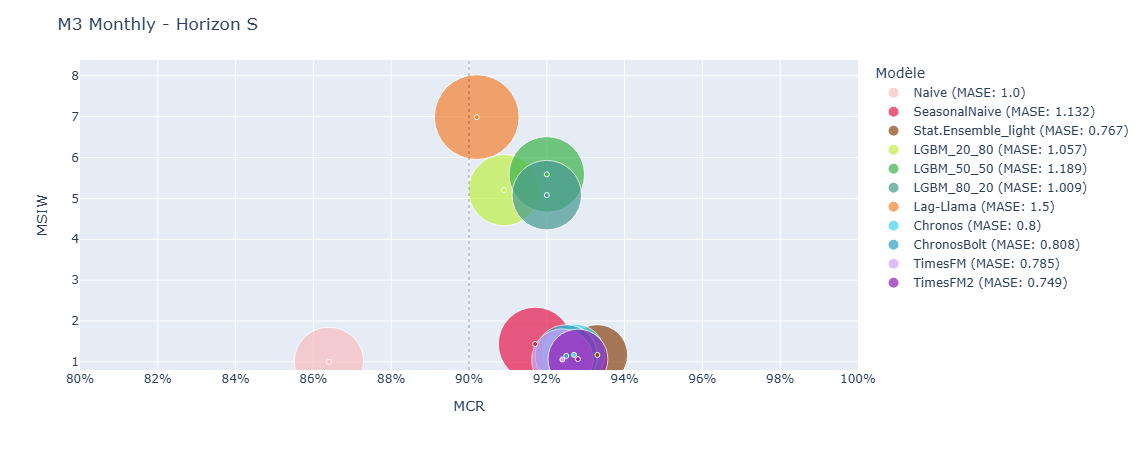}
    \includegraphics[width=\textwidth]{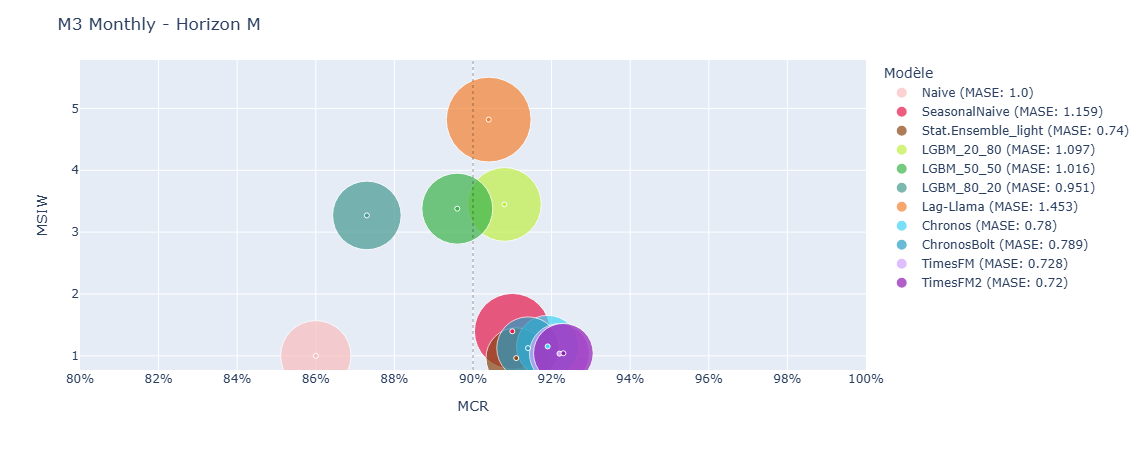}
    \includegraphics[width=\textwidth]{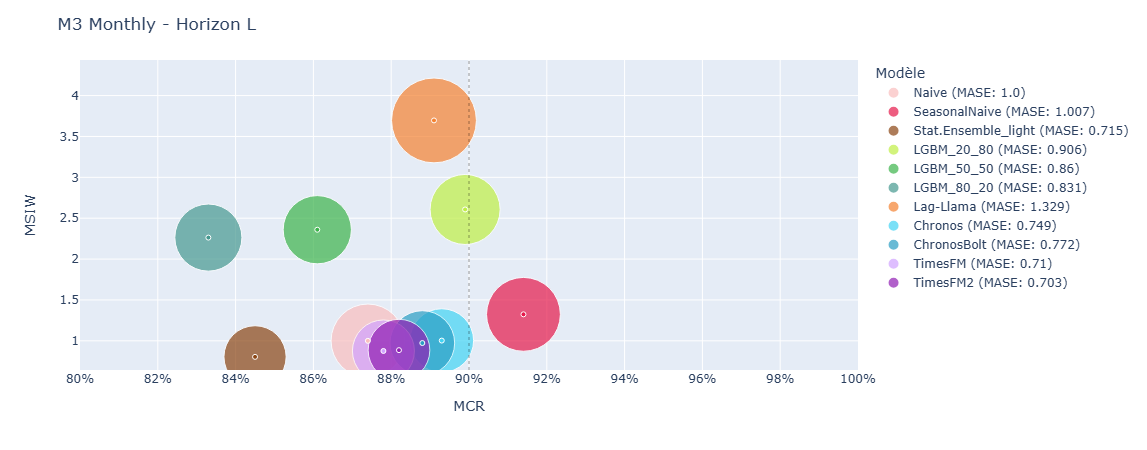}
    \caption{Split conformal prediction on M3 Monthly}
    \label{fig:m3m_result}
\end{figure}

Figures \ref{fig:nn5d_result}, \ref{fig:nn5w_result}, and \ref{fig:m3m_result} shows the results in terms of MCR, MSIW, and MASE. Overall, the different between the performance of TSFMs and other models were less pronounced than in the ERCOT case. We observed a similar correlation between the MASE and the MSIW. This can be interpreted as models with more accurate prediction also give narrower prediction intervals. 

In terms of MASE, Chronos, ChronosBolt, TimesFM, and TimesFM2 were again among the best models. For example, in the NN5-weekly-L setting, TimesFM2 achieved the best MASE (0.679) and MSIW (0.868), while respecting the target coverage (92.5\%).  

Regarding the quality of uncertainty quantification across the different approaches, it is important to note that the number of available data points is very different between different data frequency (daily and weekly for NN5, and montly for M3, see Table \ref{tab:datasets_description}). 
This impacts the models' performance in the context of conformal prediction, notably because when only a few timesteps are allocated to the calibration process, $\left| Cal \right|$ is small, hence the estimation of $1-\hat{\alpha}$ in Eq. \ref{eq:uncertainty_threshold} become imprecise.

For NN5 Daily, ChronosBolt gave the best overall results, closely followed by TimesFM2, TimesFM, Chronos, and StatsEnsemble\_light. The LGBM failed to provide the adaptive prediction interval width, likely because of the lack of training observations and exogenous variables. Lag-llama performed poorly in comparison. 
For NN5 Weekly, TimesFM and TimesFM2 gave good predictions, consistently respecting the target coverage rate with top MASE and MSIW. We observed poor performance of LGBM models, which suffered due to the limited number of data points allocated to calibration.
Regarding M3 Monthly, StatsEnsemble\_light, TimesFM, and Chronos-related models were leading, closely followed by Seasonal Naive. LGBM related models and Lag-llama presented really poor performances. Unlike in the NN5 weekly experiment, LGBM\_80\_20 exhibited similar performance to LGBM\_20\_80 and LGBM\_50\_50 in terms of MSIW. This similarity is likely due to the low cardinality of the calibration set, which prevents us from obtaining a robust picture of the regressor's errors, even though the issue of considering the maximum error value was addressed. This takes the form of reduced coverage rates.

\begin{figure}[H]
    \centering
    \includegraphics[width=\linewidth]{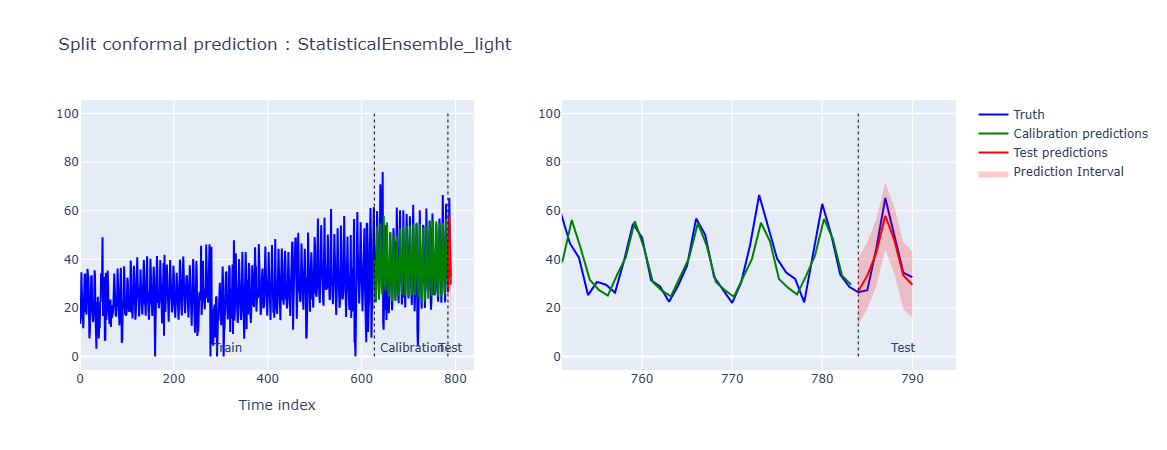}
    \includegraphics[width=\linewidth]{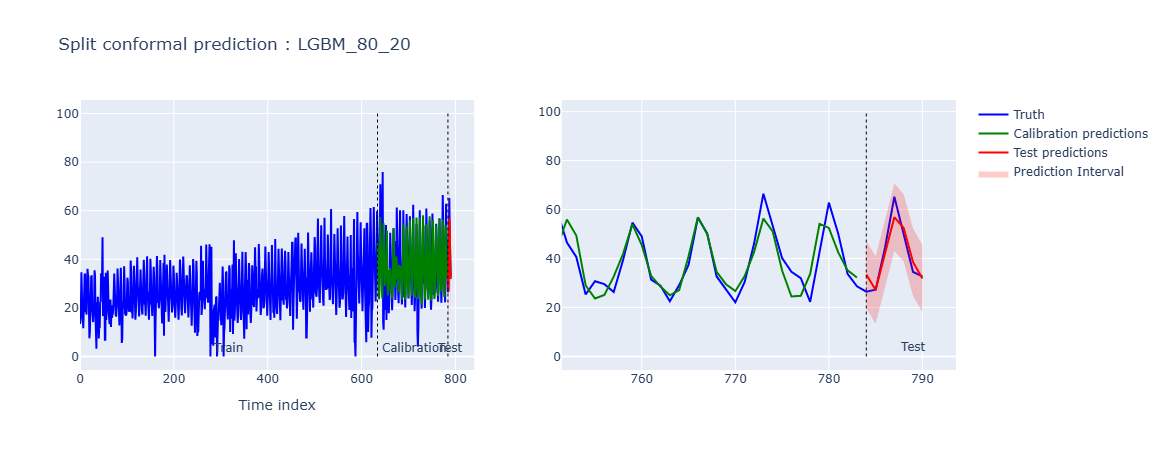}
    \includegraphics[width=\linewidth]{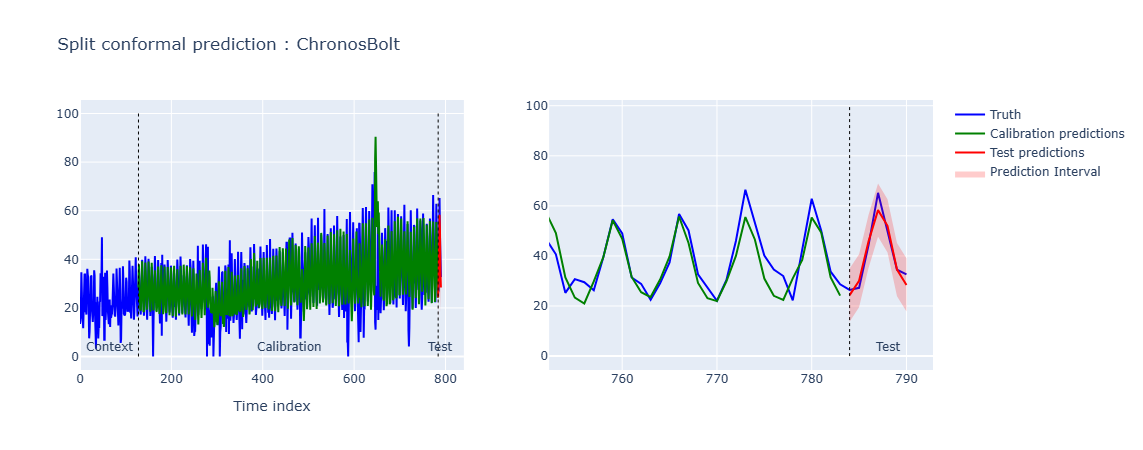}
    \includegraphics[width=\linewidth]{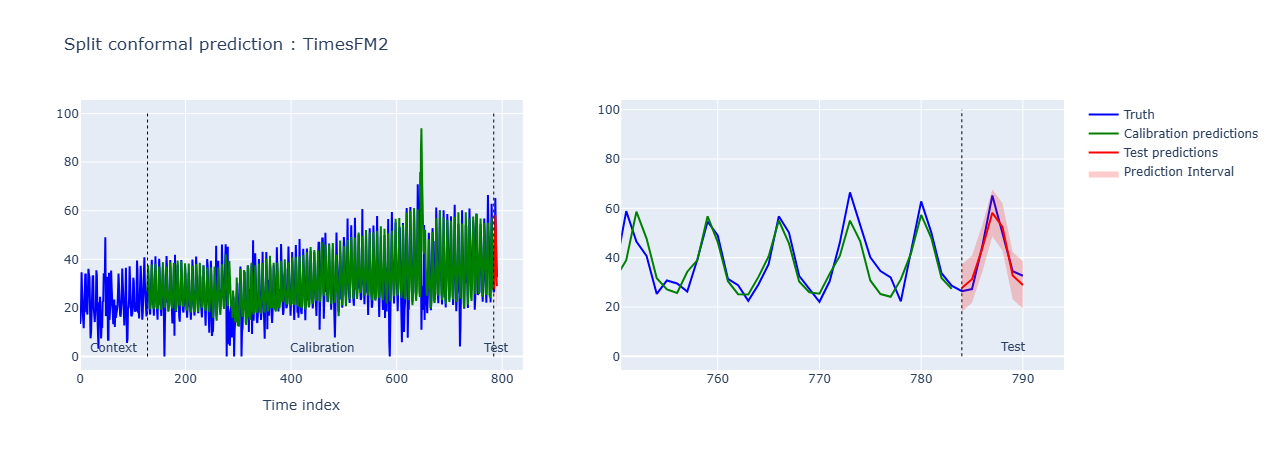}
    \caption{Examples of split conformal prediction made by different models on one time series in the NN5 Daily S setting.}
    \label{fig:res2}
\end{figure}

\begin{figure}[H]
    \centering
    \includegraphics[width=\linewidth]{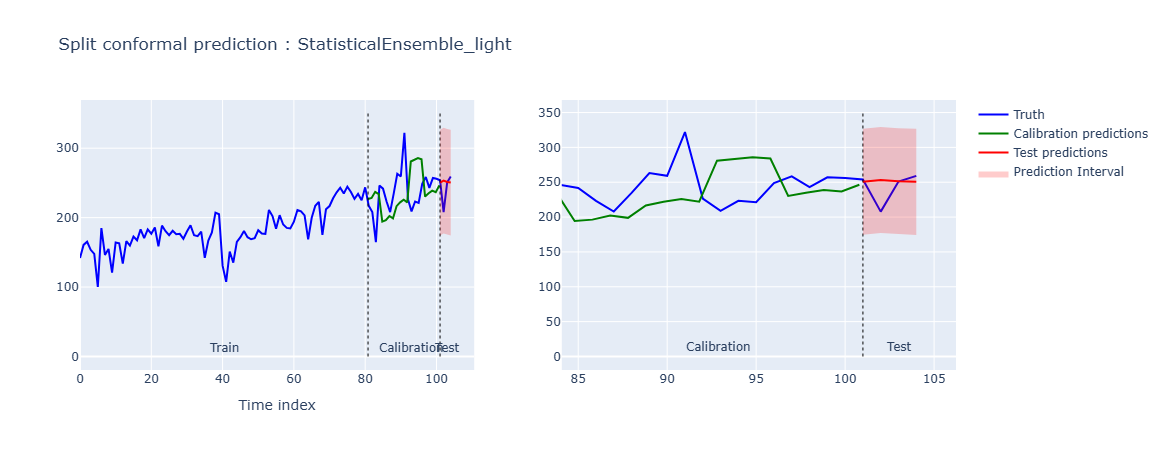}
    \includegraphics[width=\linewidth]{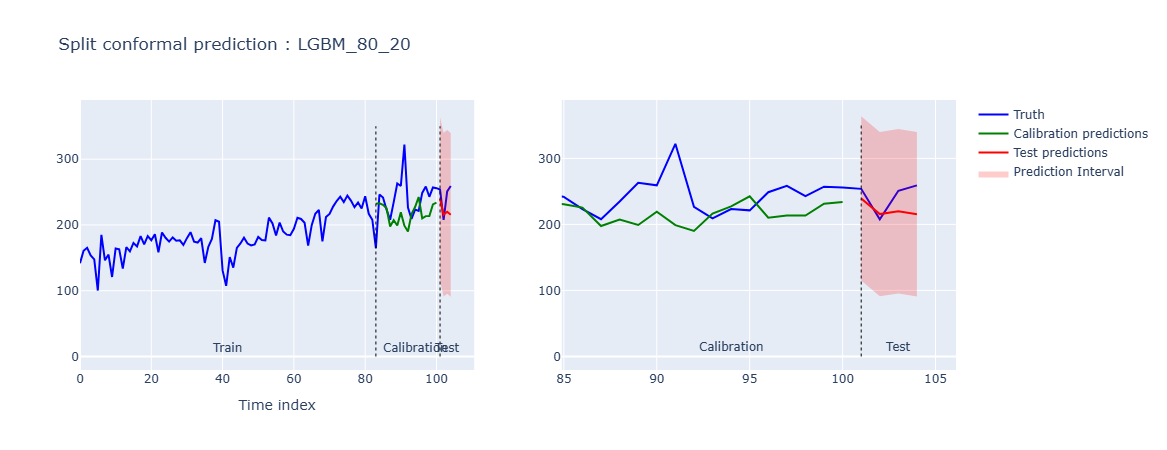}
    \includegraphics[width=\linewidth]{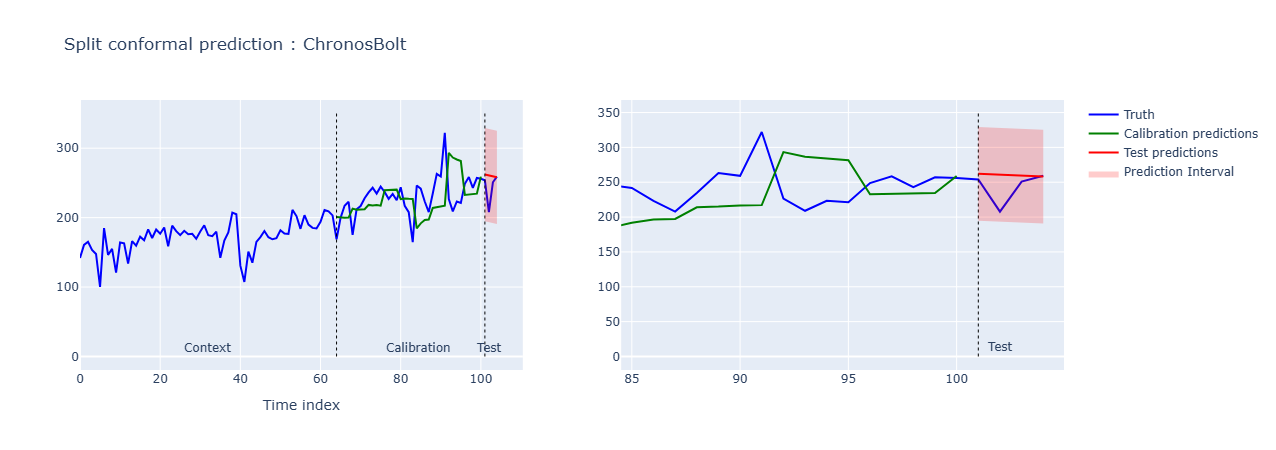}
    \includegraphics[width=\linewidth]{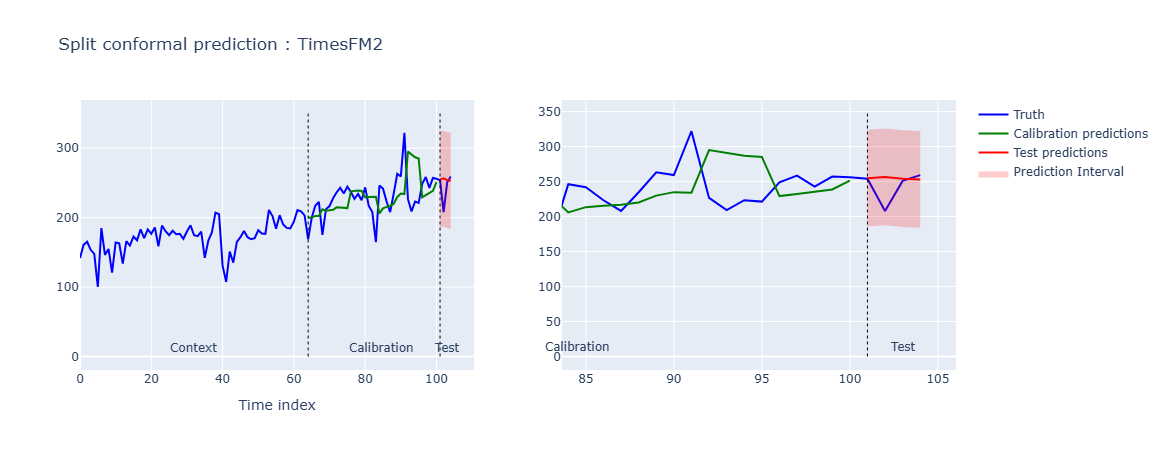}
    \caption{Examples of split conformal prediction made by different models on one time series in the NN5 Weekly S setting.}
    \label{fig:res3}
\end{figure}

\begin{figure}[H]
    \centering
    \includegraphics[width=\linewidth]{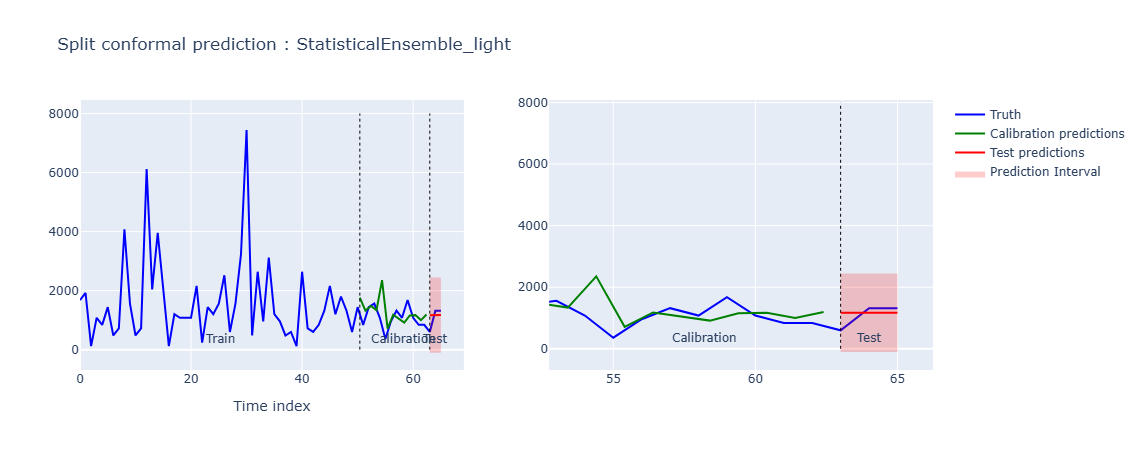}
    \includegraphics[width=\linewidth]{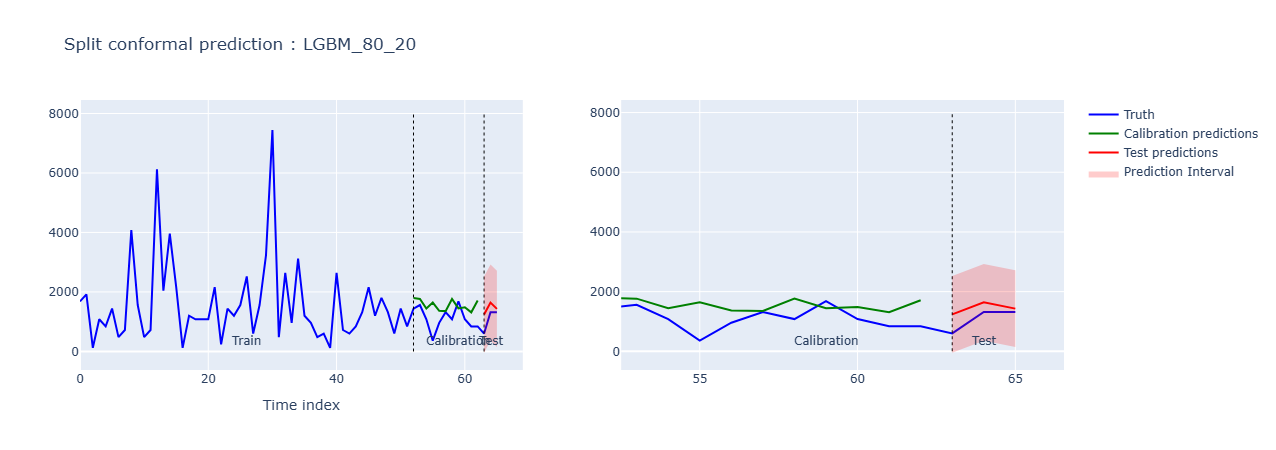}
    \includegraphics[width=\linewidth]{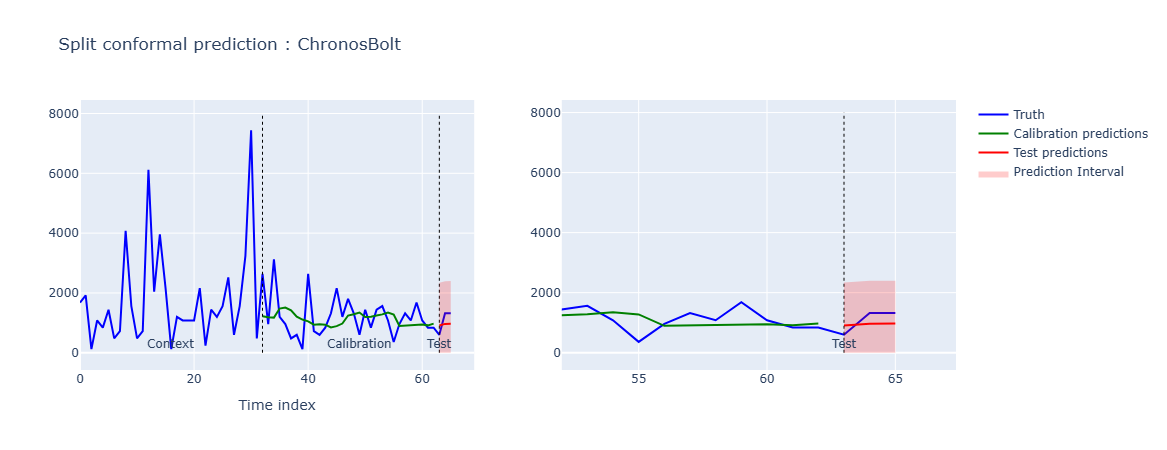}
    \includegraphics[width=\linewidth]{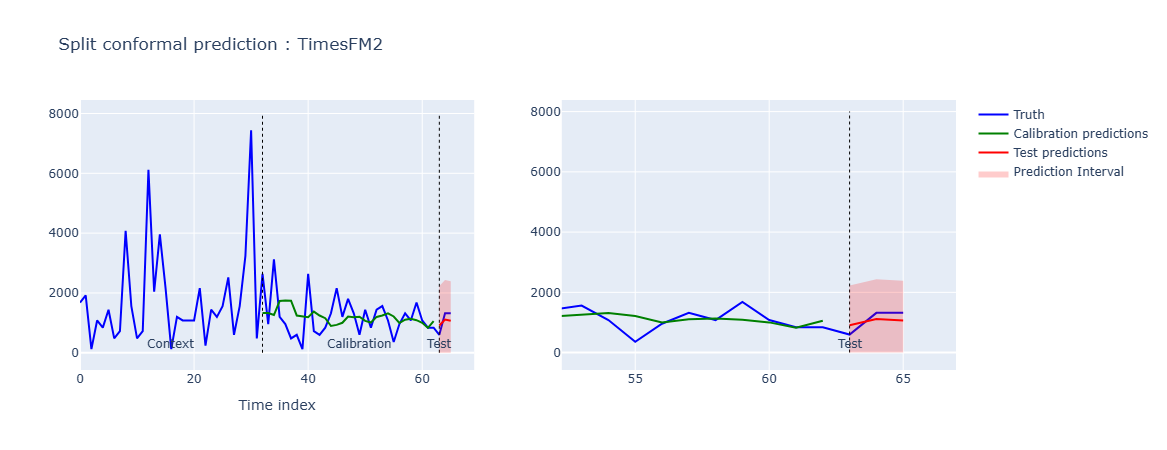}
    \caption{Examples of split conformal prediction made by different models on one time series in the M3 Monthly S setting.}
    \label{fig:res4}
\end{figure}

The complete results for the different horizons are reported in Table \ref{multivariate_results} in Appendix. Some examples of the predicted series are shown in Figures \ref{fig:res2}, \ref{fig:res3}, and \ref{fig:res4}. Most models tend to be less accurate at higher levels of granularity, as evidenced by the results from the NN5 Weekly and M3 Monthly datasets.

The conclusions remain consistent across different horizons, which enhances the robustness of the insights derived above.

\section{Related work}
\label{sec:related_work}

Recent successes of FMs in NLP, CV and multimodal \cite{zhou2024comprehensive} have catalyzed significant progress in time series forecasting through the development of time series foundation models (TSFMs). These models are designed to capture complex temporal dependencies by exploiting large-scale pre-training and transfer learning techniques, much like their counterparts in NLP, CV and multimodal. TSFM is a very active research and development domain with the introduction of seveval foundation models, such as Chronos \cite{ansari2024chronoslearninglanguagetime}, Lag-Llama \cite{rasul2024lagllamafoundationmodelsprobabilistic}, MOMENT \cite{goswami2024momentfamilyopentimeseries}, MOIRAI \cite{woo2024unified}, TimesFM \cite{das2023decoder}. In parallel, there has been a growing interest in systematically benchmarking these TSFMs to understand their strengths and limitations in diverse forecasting scenarios \cite{Nixtla2023}, \cite{aksu2024gift}.

The mentioned works focus on either developing new TSFMs, or benchmarking the forecasting capabilities of TSFMs. In this paper, we focus on another aspect of their utility: exploring an under-investigated the potential of TSFMs within the framework of conformal prediction (CP).

\section{Discussion and future work}
\label{sec:discussion}

In this paper, we evaluated the effectiveness of time series foundation models (TSFMs) within a conformal prediction framework. It should be interpreted as an introduction of TSFMs in conformal prediction. We compared the performance of different TSFMs with traditional statistical and gradient boosting methods, highlighting one main advantages of FMs in data-constrained scenarios: better prediction accuracy. These findings suggest that FMs can improve the reliability of conformal prediction in time series applications, especially when the data are scarce.

In a setup with a limited number of datasets and utilizing split conformal prediction, our results show that TSFMs, particularly TimesFM and Chronos family models, demonstrate excellent performance in both point forecasts and prediction intervals. This holds true across various data frequencies, prediction horizon lengths, and amounts of available data, outperforming traditional forecasting methods.
Their superiority is more profound when the data are very limited, because TSFMs do not need data allocated for training but only for context, which means that more data can be allocated for calibration. The trade-off between allocating data for training or calibration is more challenging for traditional methods. This phenomenon becomes even more pronounced when the creation of lags is necessary, particularly for tabular methods (LightGBM in this case). 

In terms of inference time, FMs are at least as efficient as statistical models. Our own experiments provided similar findings with that in \cite{ansari2024chronoslearninglanguagetime}.This is especially true given the numerous developments around FMs, which significantly enhance existing methods. For example, Chronos Bolt demonstrates impressively fast inference times.

Overall, we found that:

\begin{itemize}
    \item Time series foundation models are not of the same scale. Lag-Llama is the smallest with 2.5M parameters whereas TimesFM2 is the biggest with 500M parameters. The experiments show that the bigger models give better results, indicating that TSFMs are not yet at the plateau of the scaling rule. 
    \item When data are scarce, TSFMs tend to perform better than traditional method in conformal prediction settings. 
    

    \item Despite their size, TSFMs—especially optimized variants like Chronos Bolt—achieve inference speeds at least comparable to or better than traditional statistical models.
    
\end{itemize}

Our experiments still have some limitations:
\begin{itemize}
    \item We exclusively used point forecasting for all models, although some, such as Chronos models and Lag-Llama can output distribution forecast.

    \item The choice of context length may be open to criticism. TSFMs require a long context to make well-suited predictions, but we assume that the data points allocated to the context are less critical for these pre-trained models. The question of this trade-off hasn’t been addressed in details. 
    \item Split conformal prediction is not the optimal conformal prediction methods for time series. Indeed, our data do not meet the exchangeability criteria that guarantee marginal coverage. This is why some methods did not achieve the target coverage guarantee every time. To improve the study, advanced method, for example conformalized quantile regression \cite{romano2019conformalizedquantileregression}, can be used.

\end{itemize}

Although we were aware of those limitations, we chose the settings presented in this paper in favor of the computational resources required to run the experiments. 

In the near future, it would be beneficial to conduct these tests on a broader range of datasets to enhance the robustness of the results and to gauge if there are performance differences in specific application domains. Regarding TSFMs, it would be valuable to include other methods that have proven their accuracy in prediction within this benchmark. Additionally, varying the amount of data allocated to context and calibration would be interesting to quantify the difficulty presented by this trade-off more precisely. For conformal prediction, other CP methods more suited to time series could be considered.

\printbibliography
\newpage
\begin{appendices}
    \begin{flushleft}
        \Huge \textbf{Appendix}
    \end{flushleft}

    \section{Visualisation of datasets distribution}
        \begin{figure}[h!]
            \centering
            \includegraphics[width=\linewidth]{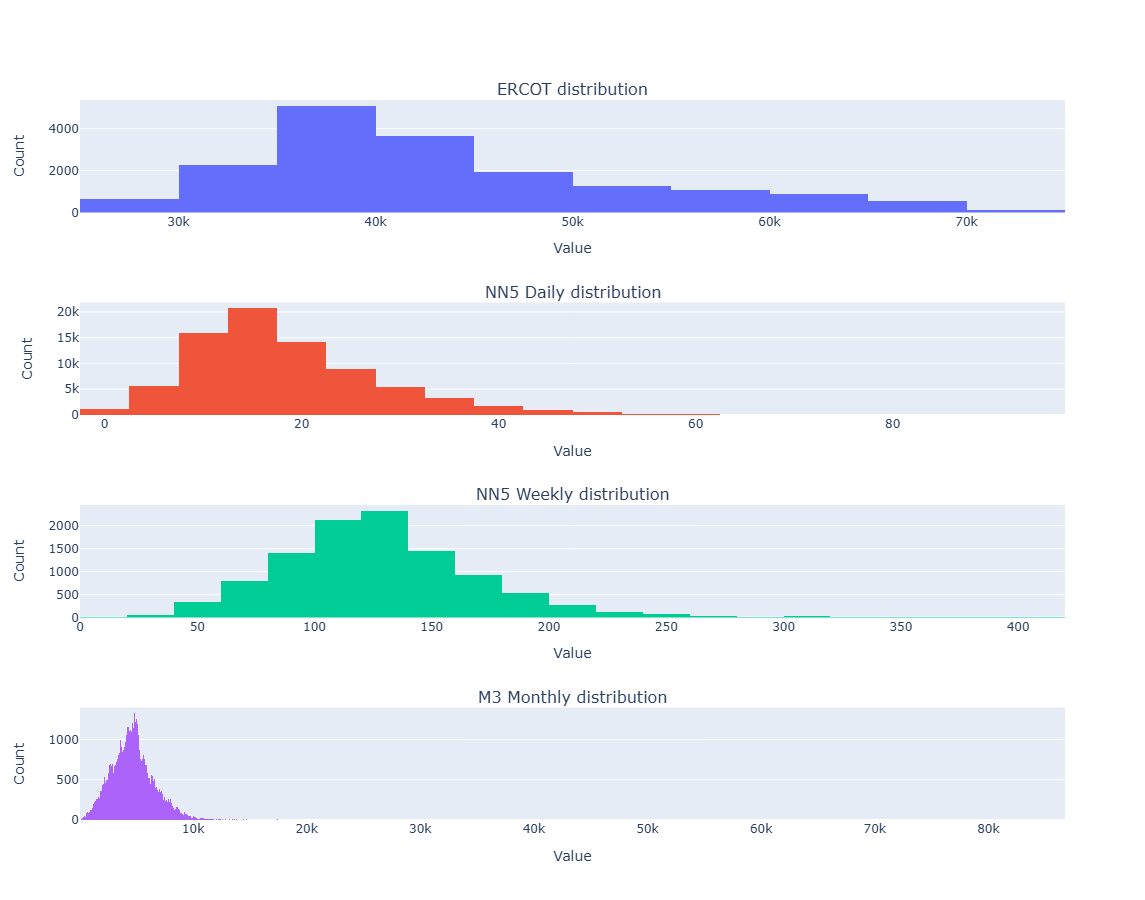}
            \caption{Datasets distribution}
        \end{figure}
    
    \section{Normalized interval width for global quantiles}

    Global quantiles are used only in the context of multivariate datasets, namely NN5 daily, NN5 weekly, and M3 Monthly. We obtain one interval width which are $l$ for a given model and $l_{naive}$ with the naive method. Then, the normalized interval width for our given model is: $$\frac{l}{l_{naive}}$$

\section{Results with local quantiles}

The results with local quantiles for \ref{subset:ercot} are presented in Table \ref{ercot_results}, while those for \ref{subsec:results_nn5_m3} are shown in Table \ref{multivariate_results}.

\begin{table}[h!]
    \centering
    \renewcommand{\arraystretch}{1.2} 
    \resizebox{\textwidth}{!}{ 
    \begin{tabular}{lccccccccc}
        \toprule
        \multirow{3}{*}{\textbf{Horizon}} & \multirow{3}{*}{\textbf{Model}} & \multicolumn{4}{c}{\textbf{ERCOT 8760 points}} & \multicolumn{4}{c}{\textbf{ERCOT 2232 points}} \\
                \cmidrule(lr){3-6} \cmidrule(lr){7-10}
                & & \textbf{MASE} & \textbf{MCR} & \textbf{IW} & \textbf{MSIW} & \textbf{MASE} & \textbf{MCR} & \textbf{IW} & \textbf{MSIW} \\
        \midrule
        \multirow{12}{*}{S} & Naive & 1.000 & 87.5 & 15946 & 1.000 & 1.000 & 87.9 & 15160 & 1.000 \\
        & SeasonalNaive & 0.795 & \underline{91.7} & 5504 & 0.461 & 1.338 & 79.2 & 6742 & 0.882 \\
        & Stat.Ensemble\_light & x & x & x & x & 1.018 & 87.9 & 6119 & 0.801 \\
        & LGBM\_20\_80 & 0.592 & \textbf{96.7} & 10748 & 0.708 & 1.195 & 89.2 & 9328 & 0.724 \\
        & LGBM\_50\_50 & 0.580 & 89.8 & 7869 & 0.511 & 0.841 & 87.9 & 7266 & 0.559 \\
        & LGBM\_80\_20 & 0.574 & 89.0 & 5621 & 0.369 & 0.659 & 89.0 & 5655 & 0.428 \\
        & Lag-Llama & 0.646 & 89.0 & 6976 & 0.458 & 0.895 & 88.7 & 7205 & 0.555 \\
        & Chronos & 0.522 & 82.7 & 4396 & 0.290 & 0.428 & 89.0 & 4206 & 0.334 \\
        & ChronosBolt & \textbf{0.430} & 86.3 & \textbf{4113} & \textbf{0.271} & \textbf{0.362} & \textbf{91.3} & \textbf{3953} & \textbf{0.312} \\
        & TimesFM & \underline{0.440} & 89.8 & 4411 & 0.291 & \underline{0.398} & \underline{90.0} & 4258 & 0.337 \\
        & TimesFM2 & 0.494 & 85.8 & \underline{4296} & \underline{0.283} & 0.416 & 89.4 & \underline{4108} & \underline{0.322} \\
        \midrule
        \multirow{12}{*}{M} & Naive & 1.000 & 86.7 & 16610 & 1.000 & 1.000 & 83.7 & 16495 & 1.000 \\
        & SeasonalNaive & 0.607 & \underline{97.2} & 9123 & 0.698 & 0.563 & \textbf{100.0} & 12457 & 1.227 \\
        & Stat.Ensemble\_light & x & x & x & x & 1.111 & 85.3 & 8002 & 0.788 \\
        & LGBM\_20\_80 & 0.763 & \textbf{98.0} & 18204 & 1.129 & 1.418 & 86.0 & 13219 & 0.888 \\
        & LGBM\_50\_50 & 0.654 & 92.7 & 13070 & 0.808 & 1.149 & 87.9 & 10763 & 0.701 \\
        & LGBM\_80\_20 & 0.607 & 90.3 & 8775 & 0.553 & 0.760 & 88.9 & 8578 & 0.552 \\
        & Lag-Llama & 0.581 & 90.6 & 8613 & 0.535 & 0.618 & \underline{96.3} & 9140 & 0.614 \\
        & Chronos & 0.604 & 81.7 & 6481 & 0.403 & 0.475 & 86.7 & 6266 & 0.415 \\
        & ChronosBolt & \underline{0.502} & 83.8 & \textbf{6006} & \textbf{0.374} & \underline{0.417} & 88.5 & \textbf{5954} & \textbf{0.395} \\
        & TimesFM & 0.525 & 87.4 & 6375 & 0.397 & \underline{0.417} & 89.0 & 6412 & 0.428 \\
        & TimesFM2 & \textbf{0.500} & 83.2 & \underline{6110} & \underline{0.380} & \textbf{0.396} & 88.5 & \underline{6070} & \underline{0.400} \\
        \midrule
        \multirow{12}{*}{L} & Naive & 1.000 & 91.8 & 17179 & 1.000 & 1.000 & 86.7 & 16486 & 1.000 \\
        & SeasonalNaive & 1.561 & 72.0 & 9527 & 0.652 & 0.932 & \textbf{100.0} & 13609 & 1.058 \\
        & Stat.Ensemble\_light & x & x & x & x & 1.371 & 81.2 & 8785 & 0.683 \\
        & LGBM\_20\_80 & 0.845 & \textbf{96.7} & 17713 & 1.034 & 1.485 & 86.0 & 12115 & 0.790 \\
        & LGBM\_50\_50 & 0.793 & \underline{92.3} & 12849 & 0.774 &1.389 & 87.2 & 10909 & 0.715 \\
        & LGBM\_80\_20 & 0.769 & 90.2 & 9410 & 0.557 & 0.923 & 88.5 & 8677 & 0.557 \\
        & Lag-Llama & 0.731 & 90.2 & 9428 & 0.562 & 0.755 & \underline{93.8} & 11128 & 0.708 \\
        & Chronos & 0.690 & 83.9 & 7750 & 0.462 & 0.580 & 91.4 & 7514 & 0.491 \\
        & ChronosBolt & \underline{0.633} & 86.3 & \textbf{7076} & \textbf{0.423} & \underline{0.526} & 90.4 & \textbf{6897} & \underline{0.451} \\
        & TimesFM & \textbf{0.599} & 86.8 & 7573 & 0.452 & \textbf{0.522} & 93.1 & 7541 & 0.493 \\
        & TimesFM2 & 0.653 & 82.8 & \underline{7095} & \underline{0.424} & 0.552 & 89.8 & \textbf{6985} & \textbf{0.449} \\
        \bottomrule
    \end{tabular}
    }
    \vspace{0.1cm}
    \caption{Results on the ERCOT dataset with 8760 and 2232 points.}
    \label{ercot_results}
\end{table}

\begin{table}[h!]
    \centering
    \renewcommand{\arraystretch}{1.2} 
    \resizebox{\textwidth}{!}{ 
    \begin{tabular}{lccccccccccccc}
        \toprule
        \multirow{3}{*}{\textbf{Horizon}} & \multirow{3}{*}{\textbf{Model}} & \multicolumn{4}{c}{\textbf{NN5 Daily}} & \multicolumn{4}{c}{\textbf{NN5 Weekly}} & \multicolumn{4}{c}{\textbf{M3 Monthly}} \\
        \cmidrule(lr){3-6} \cmidrule(lr){7-10} \cmidrule(lr){11-14}
        & & \textbf{MASE} & \textbf{MCR} & \textbf{IW} & \textbf{MSIW} & \textbf{MASE} & \textbf{MCR} & \textbf{IW} & \textbf{MSIW} & \textbf{MASE} & \textbf{MCR} & \textbf{IW} & \textbf{MSIW} \\
        \midrule
        \multirow{12}{*}{S} & Naive & 1.000 & 88.1 & 16.59 & 1.000 & 1.000 & 88.0 & \textbf{32.13} & \textbf{1.000} & 1.000 & 86.4 & 1228 & \textbf{1.000} \\
        & SeasonalNaive & 0.455 & 96.6 & 11.20 & 0.696 & 0.913 & 96.3 & 46.22 & 1.457 & 1.132 & 91.7 & 1560 & 1.436 \\
        & Stat.Ensemble\_light & 0.365 & \textbf{98.6} & 10.21 & 0.636 & \underline{0.847} & \textbf{97.3} & 49.30 & 1.573 & \underline{0.767} & \textbf{93.3} & 1215 & 1.168 \\
        & LGBM\_20\_80 & 0.440 & 96.1 & 9.54 & 0.660 & 1.101 & 95.8 & 42.16 & 1.471 & 1.057 & 90.9 & 1430 & 5.202 \\
        & LGBM\_50\_50 & 0.440 & 96.9 & 9.66 & 0.668 & 0.973 & \underline{97.0} & 42.90 & 1.499 & 1.189 & 92.0 & 1536 & 5.588 \\
        & LGBM\_80\_20 & 0.420 & 98.3 & 11.52 & 0.797 & 1.334 & 96.7 & 76.94 & 2.754 & 1.009 & 92.0 & 1397 & 5.080 \\
        & Lag-Llama & 0.749 & 93.7 & 14.55 & 0.890 & 1.055 & 94.8 & 41.78 & 1.451 & 1.500 & 90.2 & 1901 & 6.990 \\
        & Chronos & \textbf{0.336} & \underline{98.4} & 9.61 & 0.598 & 0.868 & 95.5 & 41.45 & 1.271 & 0.800 & 92.7 & 1210 & 1.171 \\
        & ChronosBolt & 0.370 & 97.7 & 8.68 & 0.543 & 0.851 & \textbf{97.3} & 43.50 & 1.343 & 0.808 & 92.5 & 1162 & 1.146 \\
        & TimesFM & \underline{0.340} & 97.7 & \underline{8.41} & \underline{0.525} & \textbf{0.826} & 96.5 & \underline{37.81} & \underline{1.207} & 0.785 & 92.4 & \textbf{1074} & \underline{1.056} \\
        & TimesFM2 & \underline{0.340} & 97.7 & \textbf{8.36} & \textbf{0.522} & \textbf{0.826} & 95.8 & 39.23 & 1.259 & \textbf{0.749} & \underline{92.8} & \underline{1080} & 1.067 \\
        \midrule
        \multirow{12}{*}{M} & Naive & 1.000 & 84.9 & 16.45 & 1.000 & 1.000 & 94.6 & 39.37 & \underline{1.000} & 1.000 & 86.0 & 1512 & \underline{1.000} \\
        & SeasonalNaive & 0.605 & 90.2 & 12.26 & 0.766 & 1.660 & 82.4 & 43.86 & 1.097 & 1.159 & 91.0 & 1839 & 1.396 \\
        & Stat.Ensemble\_light & 0.508 & 94.1 & 10.39 & 0.669 & 0.891 & 96.9 & 47.14 & 1.164 & 0.740 & 91.1 & \textbf{1266} & \textbf{0.963} \\
        & LGBM\_20\_80 & 0.505 & 93.6 & 10.53 & 0.728 & 1.221 & \underline{97.8} & 44.73 & 1.239 & 1.097 & 90.8 & 1539 & 3.448 \\
        & LGBM\_50\_50 & \underline{0.496} & \underline{94.4} & 10.52 & 0.727 & 1.243 & \underline{97.8} & 43.17 & 1.190 & 1.016 & 89.6 & 1509 & 3.379 \\
        & LGBM\_80\_20 & \textbf{0.466} & \textbf{98.0} & 14.65 & 1.012 & 1.407 & \textbf{99.9} & 71.50 & 1.987 & 0.951 & 87.3 & 1460 & 3.270 \\
        & Lag-Llama & 0.785 & 91.8 & 15.05 & 0.926 & 1.119 & 95.5 & 44.63 & 1.110 & 1.453 & 90.4 & 2142 & 4.819 \\
        & Chronos & 0.503 & 91.5 & 10.25 & 0.639 & \textbf{0.410} & 84.5 & 44.66 & 1.206 & 0.780 & 91.9 & 1467 & 1.152 \\
        & ChronosBolt & 0.506 & 90.2 & 9.02 & 0.566 & \underline{0.445} & 79.1 & 44.65 & 1.193 & 0.789 & 91.4 & 1404 & 1.124 \\
        & TimesFM & 0.519 & 88.5 & \underline{8.83} & \underline{0.553} & 0.852 & 93.8 & \underline{35.73} & \textbf{0.897} & \underline{0.728} & \underline{92.2} & \underline{1310} & 1.035 \\
        & TimesFM2 & 0.522 & 88.2 & \textbf{8.74} & \textbf{0.547} & 0.860 & 92.9 & \underline{35.37} & \textbf{0.897} & \textbf{0.720} & \textbf{92.3} & 1312 & 1.041 \\
        \midrule
        \multirow{12}{*}{L} & Naive & 1.000 & 86.0 & 17.90 & 1.000 & 1.000 & 90.2 & 42.82 & 1.000 & 1.000 & 87.4 & 1678 & 1.000 \\
        & SeasonalNaive & 0.525 & 92.7 & 12.65 & 0.725 & 1.122 & 88.4 & 46.39 & 1.109 & 1.007 & \textbf{91.4} & 1941 & 1.324 \\
        & {Stat.Ensemble\_light} & 0.430 & \underline{93.4} & 10.71 & 0.611 & 1.059 & 88.4 & 45.01 & 1.078 & 0.715 & 84.5 & \textbf{1236} & \textbf{0.804} \\
        & {LGBM\_20\_80} & 0.506 & 92.2 & 10.66 & 0.680 & 0.851 & \underline{98.7} & 45.35 & 1.198 & 0.906 & \underline{89.9} & 1523 & 2.607 \\
        & {LGBM\_50\_50} & 0.537 & 91.5 & 11.18 & 0.713 & 0.957 & 98.3 & 46.70 & 1.253 & 0.860 & 86.1 & 1378 & 2.359 \\
        & {LGBM\_80\_20} & 0.478 & \textbf{94.8} & 12.36 & 0.789 & 0.948 & \textbf{99.7} & 69.47 & 1.808 & 0.831 & 83.3 & 1322 & 2.263 \\
        & {Lag-Llama} & 0.725 & 92.4 & 15.56 & 0.880 & 0.911 & 92.8 & 49.65 & 1.116 & 1.329 & 89.1 & 2160 & 3.696 \\
        & {Chronos} &  0.491 & 91.0 & 10.69 & 0.611 & 0.931 & 82.4 & 37.96 & 0.919 & 0.749 & 89.3 & 1495 & 1.003 \\
        & {ChronosBolt} & 0.422 & 92.6 & 9.12 & 0.527 & 0.932 & 83.3 & 37.40 & 0.903 & 0.772 & 88.8 & 1425 & 0.973 \\
        & {TimesFM} & \textbf{0.413} & 92.8 & 9.10 & 0.525 & \underline{0.687} & 93.6 & \underline{36.26} & \underline{0.887} & \underline{0.710} & 87.8 & \underline{1302} & \underline{0.876} \\
        & {TimesFM2} & \underline{0.419} & 92.1 & 9.06 & 0.521 & \textbf{0.679} & 92.5 & \textbf{35.51} & \textbf{0.868} & \textbf{0.703} & 88.2 & 1307 & 0.886 \\
        \bottomrule
    \end{tabular}
    }
    \vspace{0.1cm}
    \caption{Experiment 2 : NN5 Daily, NN5 Weekly, and M3 Monthly}
    \label{multivariate_results}
\end{table}

\section{Results with global quantiles}
\label{apd:global_vs_local}

The results with global quantiles for \ref{subsec:results_nn5_m3} are shown in Table \ref{multivariate_global_results}.

\begin{table}[h!]
    \centering
    \renewcommand{\arraystretch}{1.2} 
    \resizebox{\textwidth}{!}{ 
    \begin{tabular}{lccccccccccccc}
        \toprule
        \multirow{3}{*}{\textbf{Horizon}} & \multirow{3}{*}{\textbf{Model}} & \multicolumn{4}{c}{\textbf{NN5 Daily}} & \multicolumn{4}{c}{\textbf{NN5 Weekly}} & \multicolumn{4}{c}{\textbf{M3 Monthly}} \\
        \cmidrule(lr){3-6} \cmidrule(lr){7-10} \cmidrule(lr){11-14}
        & & \textbf{MASE} & \textbf{MCR} & \textbf{IW} & \textbf{MSIW} & \textbf{MASE} & \textbf{MCR} & \textbf{IW} & \textbf{MSIW} & \textbf{MASE} & \textbf{MCR} & \textbf{IW} & \textbf{MSIW} \\
        \midrule
        \multirow{12}{*}{S} & Naive & 1.000 & 90.2 &  17.12 & 1.000 & 1.000 & 88.5 & 30.00 & 1.000 & 1.000 & 88.0 & 1240 & 1.000 \\
        & SeasonalNaive & 0.455 & 95.4 & 10.97 & 0.641 & 0.913 & 95.8 & 44.26 & 1.428 & 1.132 & 91.9 & 1800 & 1.452 \\
        & Stat.Ensemble\_light & 0.365 & 98.0 & 9.68 & 0.565 & 0.847 & 98.0 & 45.93 & 1.482 & 0.767 & 91.1 & 1178 & 0.950 \\
        & LGBM\_20\_80 & 0.440 & 95.7 & 9.63 & 0.563 & 1.101 & 96.0 & 39.81 & 1.284 & 1.057 & 91.3 & 1576 & 1.271 \\
        & LGBM\_50\_50 & 0.440 & 96.1 & 9.66 & 0.564 & 1.243 & 97.8 & 39.40 & 1.001 & 1.189 & 89.9 & 1590 & 1.282 \\
        & LGBM\_80\_20 & 0.420 & 97.9 & 11.29 & 0.660 & 1.334 & 93.3 & 45.09 & 1.455 & 1.009 & 91.5 & 1425 & 1.149 \\
        & Lag-Llama & 0.749 & 94.7 & 14.78 & 0.863 & 1.055 & 95.3 & 41.78 & 1.348 & 1.500 & 92.4 & 1903 & 1.535 \\
        & Chronos & 0.336 & 97.9 & 9.52 & 0.556 & 0.868 & 95.8 & 38.64 & 1.234 & 0.800 & 92.3 & 1287 & 1.038 \\
        & ChronosBolt & 0.370 & 96.7 & 8.69 & 0.507 & 0.851 & 96.5 & 39.82 & 1.272 & 0.808 & 91.9 & 1280 & 1.032 \\
        & TimesFM & 0.340 & 97.6 & 8.44 & 0.493 & 0.826 & 93.5 & 33.55 & 1.083 & 0.765 & 91.7 & 1216 & 0.981 \\
        & TimesFM2 & 0.340 & 97.4 & 8.35 & 0.488 & 0.826 & 93.3 & 33.64 & 1.085 & 0.749 & 92.0 & 1203 & 0.970 \\
        \midrule
        \multirow{12}{*}{M} & Naive & 1.000 & 86.8 & 16.87 & 1.000 & 1.000 & 94.9 & 39.37 & 1.000 & 1.000 & 90.4 & 1680 & 1.000 \\
        & SeasonalNaive & 0.605 & 90.3 & 12.01 & 0.712 & 1.660 & 84.1 & 45.14 & 1.147 & 1.159 & 92.3 & 2100 & 1.250 \\
        & Stat.Ensemble\_light & 0.508 & 94.3 & 10.39 & 0.616 & 0.891 & 97.3 & 43.83 & 1.113 & 0.740 & 91.4 & 1293 & 0.770 \\
        & LGBM\_20\_80 & 0.505 & 93.1 & 10.41 & 0.617 & 1.221 & 97.0 & 41.07 & 1.043 & 1.097 & 91.0 & 1822 & 1.084 \\
        & LGBM\_50\_50 & 0.496 & 93.4 & 10.52 & 0.623 & 0.973 & 95.0 & 38.25 & 1.234 & 1.016 & 90.8 & 1596 & 0.950 \\
        & LGBM\_80\_20 & 0.466 & 97.4 & 14.11 & 0.836 & 1.407 & 99.6 & 48.37 & 1.229 & 0.951 & 90.7 & 1471 & 0.876 \\
        & Lag-Llama & 0.785 & 92.2 & 15.31 & 0.907 & 1.119 & 93.3 & 44.63 & 1.134 & 1.453 & 90.8 & 2037 & 1.212 \\
        & Chronos & 0.503 & 91.6 & 10.13 & 0.601 & 0.410 & 79.4 & 36.21 & 0.964 & 0.780 & 91.7 & 1355 & 0.807 \\
        & ChronosBolt & 0.506 & 90.4 & 9.07 & 0.538 & 0.445 & 74.4 & 35.71 & 0.951 & 0.789 & 91.4 & 1404 & 0.805 \\
        & TimesFM & 0.519 & 88.5 & 8.83 & 0.553 & 0.852 & 93.8 & 35.73 & 0.897 & 0.728 & 92.2 & 1310 & 1.035 \\
        & TimesFM2 & 0.522 & 88.2 & 8.74 & 0.547 & 0.860 & 92.9 & 35.37 & 0.897 & 0.720 & 91.9 & 1312 & 1.041 \\
        \midrule
        \multirow{12}{*}{L} & Naive & 1.000 & 87.9 & 18.52 & 1.000 & 1.000 & 88.8 & 40.89 & 1.000 & 1.000 & 89.5 & 1740 & 1.000 \\
        & SeasonalNaive & 0.525 & 92.8 & 12.47 & 0.673 & 1.122 & 87.5 & 46.06 & 1.126 & 1.007 & 92.1 & 1941 & 1.172 \\
        & {Stat.Ensemble\_light} & 0.430 & 93.9 & 10.64 & 0.574 & 1.059 & 89.9 & 44.23 & 1.082 & 0.716 & 89.3 & 1292 & 0.742 \\
        & {LGBM\_20\_80} & 0.506 & 91.2 & 10.62 & 0.574 & 0.851 & 98.6 & 43.49 & 1.064 & 0.906 & 90.5 & 1678 & 0.965 \\
        & {LGBM\_50\_50} & 0.537 & 90.8 & 11.30 & 0.610 & 0.957 & 97.5 & 39.06 & 0.955 & 0.860 & 90.6 & 1509 & 0.867 \\
        & {LGBM\_80\_20} & 0.478 & 94.9 & 12.35 & 0.667 & 0.949 & 99.5 & 46.40 & 1.134 & 0.831 & 89.9 & 1384 & 0.795 \\
        & {Lag-Llama} & 0.725 & 92.7 & 15.85 & 0.856 & 0.911 & 94.3 & 49.65 & 1.214 & 1.329 & 91.3 & 2179 & 1.252 \\
        & {Chronos} & 0.491 & 90.6 & 10.57 & 0.571 & 0.931 & 82.7 & 33.95 & 0.849 & 0.749 & 91.0 & 1446 & 0.831 \\
        & {ChronosBolt} & 0.422 & 92.6 & 9.12 & 0.493 & 0.932 & 83.3 & 33.57 & 0.840 & 0.772 & 90.7 & 1449 & 0.833 \\
        & {TimesFM} & 0.519 & 89.0 & 8.90 & 0.527 & 0.687 & 93.3 & 34.85 & 0.852 & 0.710 & 90.1 & 1330 & 0.764 \\
        & {TimesFM2} & 0.419 & 92.1 & 9.09 & 0.491 & 0.679 & 92.4 & 33.64 & 0.823 & 0.703 & 90.4 & 1336 & 0.768 \\
        \bottomrule
    \end{tabular}
    }
    \vspace{0.1cm}
    \caption{Experiment 2 : NN5 Daily, NN5 Weekly, and M3 Monthly}
    \label{multivariate_global_results}
\end{table}

\end{appendices}

\end{document}